\documentclass{article}
\usepackage{M-PSI}  %

\usepackage{amsmath,amsfonts}

\usepackage{algorithmic}
\usepackage{array}

\newcolumntype{H}{>{\setbox0=\hbox\bgroup}c<{\egroup}@{}}

\usepackage{caption}
\usepackage{subcaption}

\usepackage{textcomp}
\usepackage{stfloats}
\usepackage{url}
\usepackage{verbatim}
\usepackage{graphicx}
\usepackage{balance}
\usepackage{pifont}

\usepackage{graphicx}
\usepackage{soul}
\usepackage{multirow}
\usepackage{makecell} %
\usepackage{rotating}
\usepackage{enumitem}

\usepackage{booktabs} %
\usepackage{graphicx} %
\usepackage{caption}  %
\usepackage{mathrsfs}

\usepackage{array}
\usepackage{hyperref}
\usepackage{multicol}
\usepackage{amssymb} %
\usepackage{graphicx} %
\usepackage{float} %
\usepackage{algorithm}
\usepackage{algorithmic}
\usepackage{amsmath}
\usepackage{bm} 

\setbool{twocolumns}{true}

\definecolor{mygray}{gray}{0.4}
\newcommand{\mycross}{\scalebox{0.6}{\textcolor{mygray}{\ding{55}}}}
\newcommand{\qmaxval}[1]{\underline{#1}} %

\usepackage{adjustbox}
\newenvironment{autotabular}[1]
  {\begin{adjustbox}{max width=\linewidth}\begin{tabular}{#1}}
  {\end{tabular}\end{adjustbox}}

\title{Variational Encoder–Multi-Decoder (VE-MD) for Privacy-by-functional-design (Group) Emotion Recognition}

\hypersetup{
  hidelinks,
  urlcolor=red,
  pdftitle={Variational Encoder–Multi-Decoder (VE-MD) for Privacy-by-functional-design (Group) Emotion Recognition},
  pdfauthor={Anderson Augusman, Dominique Vaufreydaz, F\'ed\'erique Letu\'e},
  pdfkeywords={Group emotion recognition, affective computing, multi-task learning, variational modeling, privacy-design.},
}

\newcommand\Affiliation[2]{%
\setbox0=\hbox{#1}%
\hspace{0.02cm}#1\hspace{-\wd0}#2
}

\makeatletter
\newcommand{\AuthorEntry}[4]{%
  \mbox{%
    \hspace{0.1cm}%
    \if\relax\detokenize{#4}\relax
      #1\Affiliation{#2}\textsuperscript{#3}%
    \else
      \href{#4}{#1\Affiliation{#2}\textsuperscript{#3,\ExternalLink}}%
    \fi
    \hspace{0.1cm}%
  }%
}
\makeatother

\author{
\AuthorEntry{Anderson Augusma}{,}{1,2}{https://augusmaa.github.io/}
\AuthorEntry{Dominique Vaufreydaz}{,}{1}{https://research.vaufreydaz.org/}
\AuthorEntry{F\'ed\'erique Letu\'e}{,}{2}{https://membres-ljk.imag.fr/Frederique.Letue/}\\
\vspace{0.5em}
\textnormal{\normalsize{
{$^1$ Univ. Grenoble Alpes, CNRS, Grenoble INP, LIG, 38000 Grenoble, France}\\[0.2em] %
{$^2$ Univ. Grenoble Alpes, CNRS, Grenoble INP, LJK, 38000 Grenoble, France}\\[-0.2em]
}}
\vspace{0.25cm}
}

\begin{document}
 
 \begin{abstract}
Group Emotion Recognition (GER) aims to infer collective affect in social environments such as classrooms, crowds, and public events.
Many existing approaches rely on explicit individual-level processing, including cropped faces, person tracking, or per-person feature extraction, which makes the analysis pipeline person-centric and raises privacy concerns in deployment scenarios where only group-level understanding is needed.
This research proposes VE-MD, a Variational Encoder--Multi-Decoder framework for group emotion recognition under a privacy-aware functional design.
Rather than providing formal anonymization or cryptographic privacy guarantees, VE-MD is designed to avoid explicit individual monitoring by constraining the model to predict only aggregate group-level affect, without identity recognition or per-person emotion outputs.
VE-MD learns a shared latent representation jointly optimized for emotion classification and internal prediction of body and facial structural representations.
Two structural decoding strategies are investigated: a transformer-based PersonQuery decoder and a dense Heatmap decoder that naturally accommodates variable group sizes.
Experiments on six in-the-wild datasets, including two GER and four Individual Emotion Recognition (IER) benchmarks, show that structural supervision consistently improves representation learning.
More importantly, the results reveal a clear distinction between GER and IER: optimizing the latent space alone is often insufficient for GER because it tends to attenuate interaction-related cues, whereas preserving explicit structural outputs improves collective affect inference.
In contrast, projected structural representations seem to act as an effective denoising bottleneck for IER.
VE-MD achieves state-of-the-art performance on GAF-3.0 (up to 90.06\%) and VGAF (82.25\% with multimodal fusion with audio).
These results show that preserving interaction-related structural information is particularly beneficial for group-level affect modeling without relying on prior individual feature extraction.
On IER datasets using multimodal fusion with audio modality, VE-MD outperforms SOTA on SamSemo (77.9\%, adding  text  modality) while achieving competitive performances on MER-MULTI (63.8\%), DFEW (70.7\%) and EngageNet (69.0).
\end{abstract}


~ %

\section{Introduction}
Group Emotion Recognition (GER) aims to infer the collective emotional state of a group. Unlike Individual Emotion Recognition (IER), GER depends not only on isolated facial cues but also on inter-person interactions, contextual dynamics, and collective behavioral patterns known as social signals~\cite{vinciarelli2009social}. However, existing GER approaches frequently rely on explicit per-person detection and prior feature extraction, thereby maintaining individual-centric processing pipelines even when the final output is group-level. In affective computing, many state-of-the-art emotion recognition pipelines rely on explicit individual-level representations, such as cropped faces, facial action units (AU), or body-based features~\cite{savchenko2022neural,sharma2021audio,Sun2020,Liu2020,cassar2024reglement}. While effective, such person-centric approaches inherently depend on prior individual-processing cues, raising concerns about biometric processing, individual monitoring, and potential misuse in surveillance contexts. These concerns are particularly relevant in collective settings such as classrooms, meetings, concerts, and public gatherings, where the objective is often to understand group-level affect rather than monitor individuals.

Individual-centric processing raises concerns about privacy, responsible use, and the risk of unnecessary person-level monitoring. Recent regulatory frameworks, including the European Union's  GDPR~\cite{albrecht2016gdpr} and later AI Act~\cite{cancela2024eu}, emphasize minimizing identity-sensitive processing and restricting high-risk emotion recognition applications.
Following these regulations, this research adopts a \emph{privacy-aware architectural design} tailored to group emotion recognition.
Rather than anonymizing visual inputs or providing cryptographic privacy guarantees as approaches from the literature, it constrains the model to operate on full video frames as input and to only produce aggregated group-level predictions.
The proposed framework avoids identity recognition, person tracking, individual embedding extraction as an explicit objective, and per-person emotion prediction. By restricting outputs to collective affect, the system avoids explicit individual monitoring while preserving socially meaningful interaction cues necessary for GER.

The main contributions of this research are:
\begin{itemize}
   \item \textit{Variational Encoder--Multi-Decoder (VE-MD)}. An architecture that learns from images a shared latent representation through multi-task learning for group-level emotion recognition under a privacy-aware functional design. The multi-task objective includes internal prediction of body and facial structural representations, while avoiding identity recognition, person tracking, explicit individual feature extraction, and per-person emotion outputs.
     \item \textit{Two alternative internal structural decoding strategies.} We investigate a query-based transformer decoder (\textit{VE-MD-PersonQuery}) and a dense heatmap decoder (\textit{VE-MD-Heatmap}) to model body and facial structural information, and analyze their respective trade-offs for variable group-size settings.
    \item \textit{Extensive multi-dataset validation and ablations on six in-the-wild datasets}. The
    dataset list includes GER but also IER datasets to evaluate the generalization capabilities of the model. Using multimodal-extensions to audio and text inputs, VE-MD demonstrates consistent improvements, achieving state-of-the-art performance on the GAF-3.0, VGAF, and SAMSEMO datasets.
    \item \textit{Insight into using internal structural representation for group affect modeling}.
    Results highlight that compressing structural representations seems to attenuate inter-person interaction information critical for Group Emotion Recognition (GER), decreasing performance. A contrario, compression seems to act as a denoiser in the case of IER datasets, improving performance metrics.

\end{itemize}

\section{Related Work}
\label{sec:relatedwork}

\subsection{Privacy-Aware Emotion Recognition}

Privacy concerns in emotion recognition have been addressed primarily through two directions: (i) \emph{input-level obfuscation} and (ii) \emph{privacy-preserving computation}. 
Input-level approaches aim to reduce identity exposure by modifying or masking sensitive regions. Zitouni et al.~\cite{zitouni2022privacy} demonstrate that masking facial regions and emphasizing contextual cues can maintain competitive affect recognition performance while limiting facial identity visibility.
In audio, anonymization techniques focus on suppressing speaker identity while preserving emotional content.
For example, Chen et al.~\cite{chen2022system} benchmark anonymization systems in the Voice Privacy Challenge, highlighting privacy–utility trade-offs, while Yao et al.~\cite{yao2024npu} and Leang et al.~\cite{leang2024exploring} propose architectures designed to conceal speaker identity while retaining paralinguistic cues.
Complementary to obfuscation, privacy-preserving computation techniques aim to protect data during inference.
Pentyala et al.~\cite{pentyala2021privacy} employ Secure Multi-Party Computation (MPC) to enable encrypted video classification without exposing raw data or model parameters.
While these approaches focus on anonymizing inputs or securing computation, they do not explicitly address the functional design of group-level affect modeling systems.
This research adopts a different perspective: rather than modifying the input signal or encrypting computation, it constrains the model’s representational objectives and outputs. VE-MD is designed to avoid identity recognition, tracking, and per-person emotion prediction, producing only aggregate group-level outputs.

\subsection{Group Emotion Recognition from Individual Cues}

Most existing GER pipelines rely on prior individual processing in their representations.
Early works combine individual facial features with global scene context.
Tan et al.~\cite{tan2017group} fuse facial CNN features with scene-level representations, while Yan et al.~\cite{yan2016multi} integrate facial texture, landmarks, and audio cues via hybrid CNN–BRNN frameworks.
Graph-based approaches further model explicit inter-person relationships.
Ngoc et al.~\cite{ngoc2020facial} and Guo et al.~\cite{guo2020graph} construct graphs over facial landmarks or skeleton features to capture spatial dependencies, and Jung et al.~\cite{jung2015joint} jointly model temporal appearance and landmark geometry.
More recently, Kumar et al.~\cite{kumar2025fusing} integrate pose estimation with multimodal audio–video features for GER.
Although these methods achieve strong performance, they typically rely on explicit per-person detection, landmark extraction in advance, or prior individual-aligned features for aggregation.
Consequently, per-person modeling remains embedded in the processing pipeline, even when the final output is group-level.
In contrast, the proposed approach avoids treating structural cues as primary per-person inputs to the emotion classifier. Instead, VE-MD uses internal structural representation as auxiliary supervision to shape the latent space,
allowing group-level modeling independent of prior individual processing, while preserving interaction cues essential for GER.

\subsection{Multitask Learning for Emotion Recognition}

Multitask learning (MTL) has been widely adopted in emotion recognition to leverage shared representations across related tasks. Foggia et al.~\cite{foggia2023multi} employ hard parameter sharing to jointly predict emotion, gender, and age, while Hu et al.~\cite{hu2018deep} explore soft sharing and adversarial alignment to balance shared and task-specific features. Auxiliary structural supervision has also been explored to enhance emotion modeling. Yin et al.~\cite{yin2017multi} and Pons et al.~\cite{pons2020multitask} incorporate pose and action-unit detection as complementary tasks. HyperFace~\cite{ranjan2017hyperface} jointly learns facial detection, landmark localization, and pose estimation, and Hong et al.~\cite{hong2018multimodal} integrate manifold regularization to improve structural consistency. However, prior multitask frameworks primarily aim to improve predictive performance and do not explicitly examine how structural supervision influences collective versus individual affect modeling.
VE-MD extends MTL by introducing multi-task learning of internal structural representation on body and face skeletons, proposing two flavors of the model based on PETR-like~\cite{shi2022end} or Openpose-like~\cite{cao2019openpose} representations.

\section{Model Architecture} %
\begin{figure*}[ht]
  \centering
\includegraphics[width=0.98\linewidth]{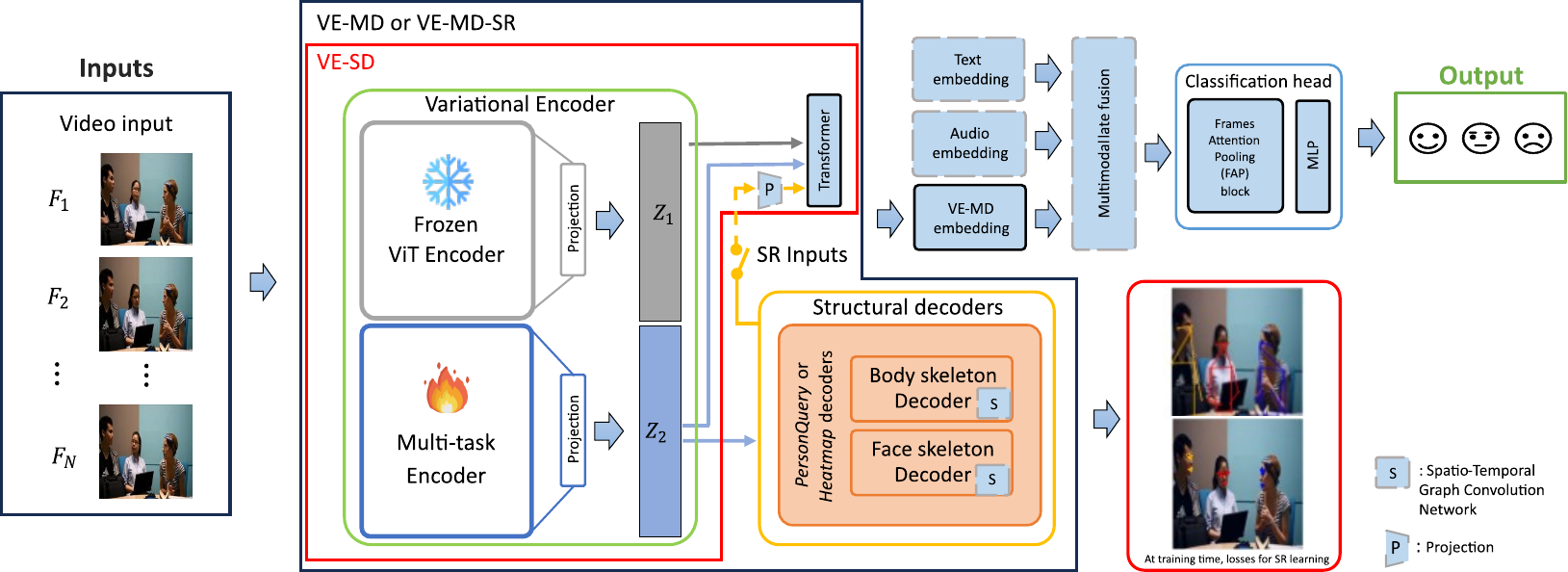}
\caption{\textit{Overview of the proposed VE-MD architecture. Left: input data, consisting of video frames or a single image. Middle: the Variational Encoder (VE, green box), which learns two latent spaces: $Z_1$ for emotion recognition and $Z_2$ for joint optimization of emotion recognition and person structural representation. Right: the multi-decoder head, where the Emotion Decoder uses $Z_1$ and $Z_2$, optionally complemented by structural outputs from the body and face decoders (SR inputs).} Optional modules are outlined with a dashed line.}
\label{fig:ve_md}
\end{figure*}

\label{sec:methodology}
The VE-MD architecture is depicted in Figure~\ref{fig:ve_md}.
It is composed of a Variational Encoder using two sub-encoders taking entire video frames as input, a frozen ViT, and a trainable multitask encoder.
This variational encoder provides the latent space representation of input video frames.
Then, a structural decoder designed to provide insights about body language and facial expression is added to the model.
In this paper, two different versions of the structural decoders are proposed and compared.
The last component is the emotion decoder, taking inputs from encoders to generate the final classification of the video clip.
During training time, internal structural information is added to guide the latent representation of the multitask encoder toward body language information.
The following sections describe these components.

\subsection{Variational Encoder for Multitask Latent Space}
\label{sec:encoders}

The VE-MD employs a  Variational Encoder (VE) design consisting of two encoder branches that create a compound latent-space for multitask learning. 
The variational encoder choice and design come from a series of experiments on end-to-end video emotion recognition presented later in sections~\ref{sec:ve_experiments}.
The first branch of the VE is a frozen pretrained version {ViT-Large} encoder~\cite{dosovitskiy2020vit}. %
The second branch is a trainable {multitask encoder}, responsible for learning meaningful representations for emotion classification and structural representation estimation (body and face).
After several experiments (see section~\ref{sec:vit_influence}), the selected multitask encoder is a ResNet-50 architecture. Both sub-encoder branches are projected to a lower dimension using a linear layer and provide $Z_1$ and $Z_2$, the two components of the encoder latent-space representation for emotion recognition in the VE-MD architecture. The latent-space representation is trained to follow a Gaussian distribution space using variational inference.

\begin{figure}[ht]
  \centering
  \includegraphics[width=0.9\linewidth]{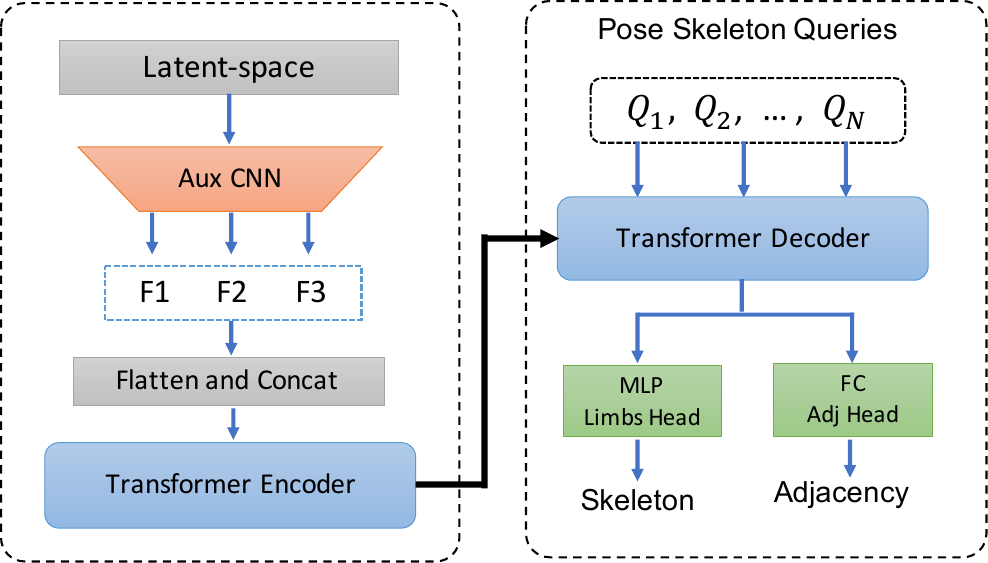}
\caption{\textit{PersonQuery} decoder. Left: the latent space is processed by an auxiliary convolutional module that produces three multi-scale feature tensors, \(\mathbf{F}_1\), \(\mathbf{F}_2\), and \(\mathbf{F}_3\), which are flattened and fed to the transformer encoder. Right: the transformer decoder receives target queries together with the encoded features and predicts the structural representation and the adjacency matrix through a multi-layer perceptron head and a fully connected head, respectively.}
  \label{fig:structural representation_decoder}
\end{figure}

\subsection{Structural Decoders for bodies and faces}

The structural decoder architecture consists of two sub-decoders: a decoder for body structural representation estimation and a decoder for facial structural representation estimation.
The two structural representation decoders receive as inputs the latent-space embedding from the multitask encoder.
Two approaches are proposed for these structural representation decoders: \textit{PersonQuery} and \textit{Heatmap}.  These propositions are detailed in the next sections.

\subsubsection{PersonQuery Decoder}

To perform the structural representation of bodies and/or faces, the \emph{PersonQuery} network follows the PETR \cite{shi2022end} style, which is based on DETR \cite{carion2020end} for fully end-to-end pose prediction.
As depicted in Figure~\ref{fig:structural representation_decoder}, the proposed structural representation decoder builds upon a transformer-based architecture to predict a limb head and an adjacency matrix.
The network takes the latent-space from the variational encoder as input.
Then, it applies an auxiliary residual convolution module (as defined in Section~\ref{sec:residual_block}) to produce three high-level features to feed the transformer module.

A learnable query embedding ($Q$) of dimension \(D\) is interpreted as a “prototype” that predicts a full structural representation for one person (a set of limbs). These queries, organized into a target sequence \(\mathbf{T}\in \mathbb{R}^{Q \times D}\) (with \(Q = \texttt{num\_queries}\)), are passed into the transformer decoder alongside the encoded source sequence from the transformer encoder.

In contrast to PETR,  the network comprises two prediction heads.
A \emph{Limb-Head}, which is a multi-layer perceptron (MLP) ended with a sigmoid, outputs line-segment endpoints for each limb.
Specifically, for \(\texttt{num\_limbs}\), it predicts \(4~\times \texttt{num\_limbs}\) coordinates, denoted \([\mathrm{x}_1, \mathrm{y}_1, \mathrm{x}_2, \mathrm{y}_2, \ldots]\) for each query. This format encodes the 2D endpoints of each limb $\mathbf{L}_{\mathrm{pred}} \in \mathbb{R}^{Q \times 4\, \times \texttt{num\_limbs}}$. 
An \emph{Adjacency-Head} which consists of a fully connected layer ended with sigmoid, output dimension~\(\texttt{num\_limbs}~\times~\texttt{num\_limbs}\) adjacency matrix for each query. This matrix specifies the connectivity scores between all pairs of limbs. Formally, $ \mathbf{A}_{\mathrm{pred}} \;\in\; \mathbb{R}^{Q \times \texttt{num\_limbs} \times \texttt{num\_limbs}}$.
Hence, each query in the decoder simultaneously outputs both limb endpoints \(\mathbf{L}_{\mathrm{pred}}\) and the joint adjacency matrix \(\mathbf{A}_{\mathrm{pred}}\).
The intuitive goal is for each query to specialize in decoding a consistent subset of structural representation connections or relationships.
By predicting the adjacency matrix, the model can capture pairwise joint dependencies of limbs, enhancing the structural representation organization~\cite{abedi2024engagement}.
To better exploit the structural dependencies of the human body, a Spatio-Temporal Graph Convolutional Network (ST-GCN)~\cite{wang2022skeleton} is inserted on top of the SR features.
Each joint is treated as a graph node, and the connections between limbs define the adjacency matrix.
The graph convolution propagates information between connected limbs (spatial modeling), while a temporal convolution aggregates information across consecutive frames (temporal modeling).

In this \textit{PersonQuery} approach, as in PETR or DETR, the model’s capacity is constrained by a fixed number of queries defined at training time.
Each query corresponds to a person to detect.
Setting the number of queries to $N$ allows the model to predict exactly $N$ persons even if more or fewer are present in the image. %
The fixed-query mechanism inherently restricts scalability and flexibility when dealing with in-the-wild group scenes containing a variable number of people.
Indeed, increasing the value of $N$ increases the number of parameters in the model in return.
Two main strategies can be used for defining this value.
$N$ can be set to an arbitrary value.
In this case, the \textit{PersonQuery} decoder uses learnable queries of size $N\times{}D$, denoted $Q_N$ in later experiments.
$N$ can be set to the maximum value of detected persons to minimize false negatives, i.e., actual persons within the image that are not detected.
This condition is named $Q_{Max}$.

\subsubsection{Heatmap decoder} 
\label{sec:ve-md-heatmap}

\begin{figure}[t]
  \centering
  \includegraphics[width=0.99\linewidth]{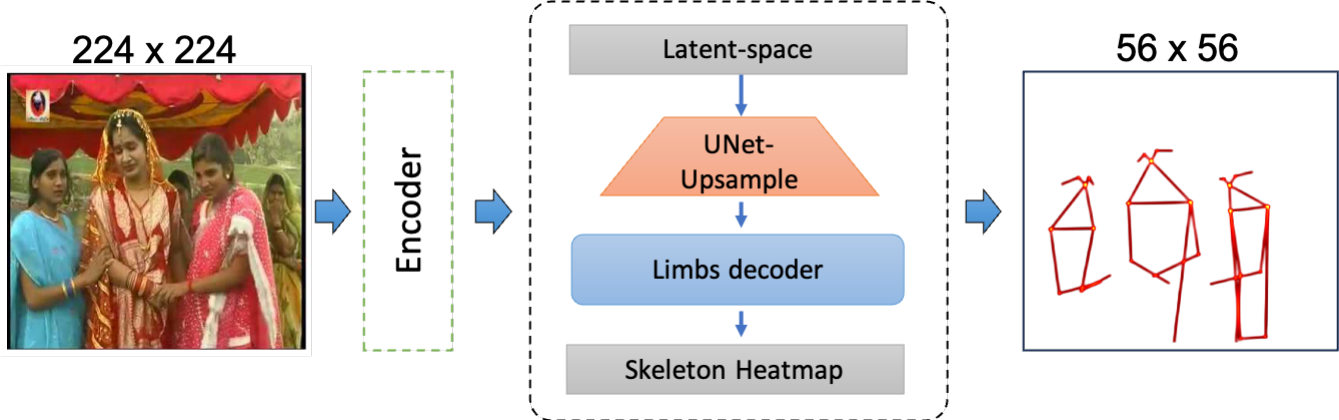}
 \caption{\textit{Heatmap} decoder. At the top is the VE latent-space input, followed by a custom UNet-upsample network, and the limbs decoder is applied to predict the limb heatmap. The same process is applied to the structural representation of faces.}
  \label{fig:SR_decoder_heatmap}
\end{figure}

To overcome the limitation of the former decoder, another proposal, called \emph{Heatmap}, operates an end-to-end estimation of direct structural representations through limb-connection estimation inspired by OpenPose~\cite{cao2019openpose}.
In contrast to \textit{PersonQuery}, the \emph{Heatmap} decoder naturally adapts to any number of individuals present in an image, since person structures emerge directly from the spatial activation patterns of the predicted heatmaps.
In literature, most structural representation or pose-estimation methods~\cite{cao2017realtime, cao2019openpose, osokin2019real, jo2022comparative, zhang2025two} rely on keypoint detection followed by a separate post-processing step to infer the structural connections.
In the \emph{Heatmap} decoder, this two-step procedure is replaced by a single-stage prediction: rather than estimating discrete keypoints and linking them afterward, the model directly predicts limb-connection heatmaps representing the structural configurations of both bodies and faces (see Figure~\ref{fig:SR_decoder_heatmap}).

To do so, processing starts by upsampling features from the VE latent-space by using a custom U-Net network~\cite{ronneberger2015u}.
Then, it outputs a heatmap of $n$ channels, $n$ assigned to the number of limbs.
Original OpenPose predicts both confidence maps and Part Affinity Fields (PAFs), i.e., 2D vectors to corresponding joints, to explicitly associate body parts to individual instances.
\textit{Heatmap} predicts only dense limb heatmaps without performing person-wise association.
This design allows structural information to be embedded directly within the latent-space, enhancing its semantic richness for emotion recognition while maintaining an end-to-end differentiable training pipeline.
By avoiding explicit part grouping or instance-level skeleton reconstruction, the decoder does not produce per-person structural outputs.
Instead, it generates a dense relational activation map that encodes structural patterns across the scene.

\subsection{Emotion Decoder}
\label{sec:emotion_decoder}
The Emotion Decoder (ED)
receives embeddings from all upstream encoders, including structural representation-based information derived from SR decoders.

\subsubsection{Standard computation schema}

For each frame, the variational encoder produces flattened features: $\mathbf{Z}_1, \mathbf{Z}_2 \in \mathbb{R}^{T\times (C_z \cdot h_z  \cdot w_z)}$,
where $(C_z, h_z, w_z)$ denote the channel and spatial dimensions of each encoder's latent-space and $T$ denotes the sequence of frames.
The concatenation of $Z_1$ and $Z_2$ provides the latent representation of the VE, defined as
$\tilde{\mathbf{Z}}_t = \phi\!\left([\mathbf{Z}_{1,t}; \mathbf{Z}_{2,t}]\right)
\in \mathbb{R}^{C}, $ where $C = 2C_z,$ and $\phi$ a linear projection.
Structural representations are then directly concatenated as
    $\mathbf{x}_t = [\tilde{\mathbf{Z}}_t; \mathbf{S}^{\text{body}}_t; \mathbf{S}^{\text{face}}_t]$, with:
    \[
    \mathbf{S}^{\text{body}},\ \mathbf{S}^{\text{face}} \in
    \begin{cases}
    \mathbb{R}^{T\times (4 N_i Q)}, & \text{for \textit{PersonQuery}},\\[4pt]
    \mathbb{R}^{T\times (H_S \cdot W_S)}, & \text{for \textit{Heatmap}}.
    \end{cases}
    \]

    where $N_i$ is the number of limbs (body or face), $Q$ is the number of detection queries, and $(H_S, W_S)$ are the spatial dimensions of the heatmaps. Consequently $ x_t \in \mathbb{R}^{D}$ with $D~=~C~+~\Delta_S$. If there is only one structural representation $\Delta_S =4Q N_i $ or $\Delta_S =H_S W_S $. If there are two (body, face), $\Delta_S =4Q(N_b + N_f)$ or $\Delta_S =2H_S W_S $.

To handle the temporal aspect, i.e., to create a rich joint embedding at video-level, a first module integrates a transformer encoder coupled with a Frames Attention Pooling (FAP) computation adapted from Okabe et al~\cite{okabe2018attentive} proposal.
FAP models temporal dependencies across frames using a formula in the following equation:

\begin{equation}
\label{fap_eq}
\mathrm{FAP} = \sum_{i=1}^{N} \alpha_i f_i
\end{equation}
where $\alpha_i=\mathrm{softmax}(s_i)$ and $s_i=w^{T}f_i+b$.

Last, the emotion decoder employs an MLP to make the final classification from the video-level embedding coming from the Frame Attention Pooling.

\subsubsection{Optional computations}

As depicted in Figure~\ref{fig:ve_md}, some modules can be optionally activated in the emotion decoder.
The first module consists of a projection dedicated to reducing the embedding size of the structural representations coming from structural decoders:
\[
\hat{\mathbf{S}}^{\text{body}}_t = \rho(\mathbf{S}^{\text{body}}_t),\quad
\hat{\mathbf{S}}^{\text{face}}_t = \rho(\mathbf{S}^{\text{face}}_t),
\]
yielding $
\mathbf{x}_t = [\tilde{\mathbf{Z}}_t; \hat{\mathbf{S}}^{\text{body}}_t ; \hat{\mathbf{S}}^{\text{face}}_t] \in \mathbb{R}^{D},$ where  $ \rho$ is the projection linear function. In this case, $\Delta_S = C_S$ for only one structural representation and $\Delta_S = 2C_S$ if there are two, with $C_S$ being the projected dimension. The effect of the projection module is questioned in section~\ref{sec:projection_effect}.

\subsection{Loss Function}
\label{sec:loss}

The overall loss function for our model integrates multiple components, each tailored for specific tasks:

\begin{itemize}[leftmargin=*]
    \item {Emotion Classification Loss:} A standard cross-entropy loss (\(\mathcal{L}_{\mathrm{cls}}\)) is used for emotion classification.
    \item Structural representation estimation Loss (\(\mathcal{L}_{\mathrm{p_{i}}}\)) for body and face. In the case of \textit{Heatmap} decoder, we use \textit{MSE-loss}, and for \emph{PersonQuery}, the structural loss is composed of limb and adjacency losses. For the limb-loss, we use \textit{smooth-L1 loss}, and for adjacency-loss, we use  \textit{binary cross entropy}.
    \item {Maximum Mean Discrepancy (MMD) Loss~\cite{dziugaite2015training}:} A regularization loss to ensure Gaussian feature representation.
\end{itemize}

The total loss is defined:

\begin{equation}
\label{eq:total_loss}
\mathcal{L}=\mathcal{L}_\mathrm{cls} + \beta_{p_1}\mathcal{L}_{p_1} + \beta_{p_2}\mathcal{L}_{p_2} + \beta_\mathrm{mmd}\mathcal{L}_\mathrm{MMD}
\end{equation}

where: %

\begin{equation}
\mathcal{L}_{\mathrm{p_{i}}} = \beta_{\mathrm{limb}}\mathcal{L}_{\mathrm{limb}} + \beta_{\mathrm{adj}}\mathcal{L}_{\mathrm{adj}}, \quad i \in \{1, 2\}
\end{equation}

with \(\beta_{\mathrm{p_{i}}}\), \(\beta_{\mathrm{limb}}\), \(\beta_{\mathrm{adj}}\), and \(\beta_{\mathrm{mmd}}\) representing loss weighting factors.

\section{Experiments on GER/IER Datasets}
\label{sec:ve_experiments}

This section presents the experimental evaluation of VE-MD in the video-only setting on both GER and IER datasets. We first describe the datasets, the automatic structural annotations, and the preliminary experiments used to define the encoder training strategy and the multi-task backbone. We then detail the training setup and report the main video-only results of the proposed VE-MD-SR variants. Multimodal extensions with audio and text are presented later in Section~\ref{sec:multimodal}.
\subsection{Datasets and Automatic Annotations}

\label{sec:datasets_annotation}
\begin{figure*}%
  \centering
  \includegraphics[width=0.97\linewidth]{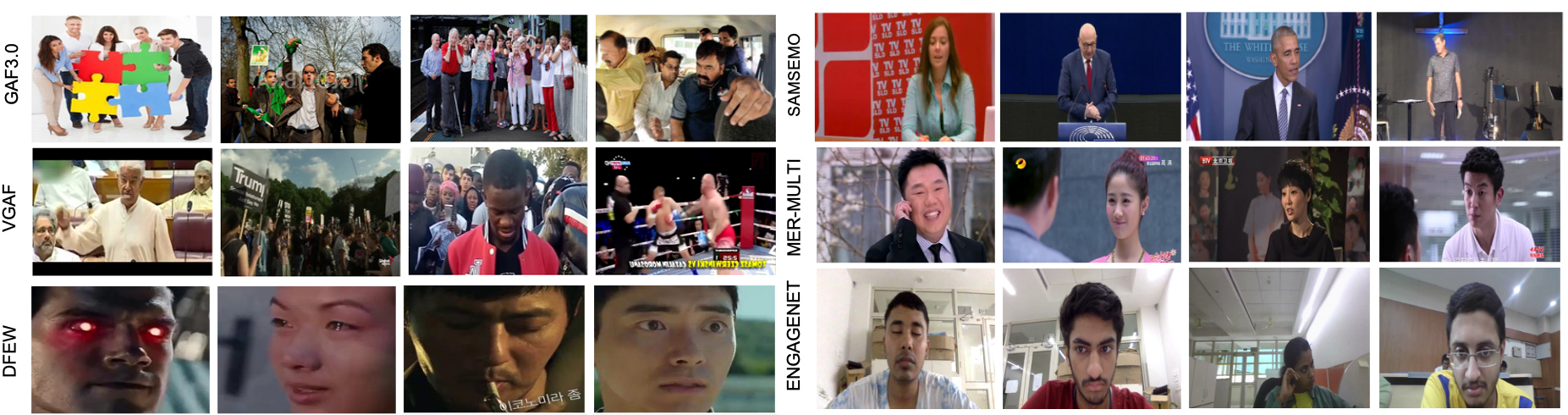}
  \caption{Overview of datasets used in this study. 
  GAF-3.0 and VGAF at left illustrate Group Emotion Recognition (GER) datasets, followed by examples from IER datasets: DFEW, SAMSEMO, MER-MULTI, and EngageNet.}
  \label{fig:datasets_used_overview}
\end{figure*}

Six \textit{in-the-wild} datasets were used for evaluation, spanning both  {Group Emotion Recognition (GER)} as our target task, and {IER} scenarios 
as a generalization task (Figure~\ref{fig:datasets_used_overview}). GER datasets are corpora where the emotion label focuses on at least two persons.
IER datasets are corpora where the emotion focuses on one individual, even though there could be other people around in the environment.

{GAF-3.0}~\cite{gupta2018attention} and  {VGAF}~\cite{dhall2020emotiw} serve as GER benchmarks, each labeled into three affective classes (\emph{Positive}, \emph{Neutral}, \emph{Negative}).  
GAF-3.0 contains diverse social contexts (meetings, sports, protests), while VGAF provides video clips with dynamic camera motion and acoustic variations.
The IER datasets include  {SAMSEMO}~\cite{bujnowski2024samsemo},  {MER-MULTI}~\cite{lian2023mer},  {DFEW}~\cite{jiang2020dfew}, and  {EngageNet}~\cite{emotiw2023}.  
SAMSEMO is a multilingual multimodal corpus (text, audio, video) annotated for six Ekman emotions plus \emph{Neutral/Other}.  
MER-MULTI consists of movies and TV shows, annotated with discrete and continuous valence measures.  
DFEW focuses on facial expression dynamics across unconstrained scenes, while EngageNet targets engagement recognition in online video-conference classrooms with four engagement levels: Not-engaged, engaged, barely-engaged, and highly-engaged.

\begin{figure}%
  \centering
  \includegraphics[width=0.98\linewidth]{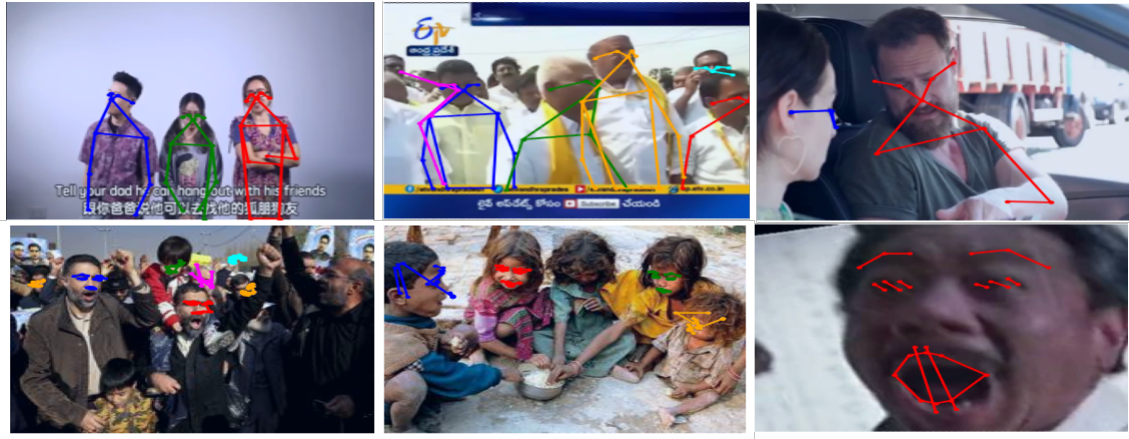}
  \caption{Example of automatic data annotation for structural representation. The first line corresponds to body annotations using ViTPose with 18 limb connections (COCO-style). The second line corresponds to face annotations using FaceAlignment with 20 custom limb connections.}
  \label{fig:data_annotation}
\end{figure}

To support multitask learning, all datasets are extended with automatic body and face structural representation annotations.
Examples of these annotations are illustrated in Figure~\ref{fig:data_annotation}.
Body pose keypoints were generated using \textit{ViTPose}~\cite{xu2022vitpose} following the COCO 17-keypoint format (to build 18 limb connections). 
Facial landmarks were extracted using \textit{FaceAlignment}~\cite{bulat2017far} and reduced from 63 to 20 connections, focusing on dynamic regions such as the lips, eyebrows, and eyes.

This automatic annotation process is applied to the training set of all datasets.
It is obviously not flawless.
In the GER datasets, when the scene is crowded, or when framing, distance, and lighting conditions are not adequate, automatic annotation can generate false negatives. Conversely, in the IER datasets, the emotion label is focused on a single person, but the automatic detection model also detects people present around them.
This is the case for SAMSEMO and MER-MULTI, where some people give lectures in front of an audience. This quality of automatic labelling has an impact on the size of the datasets used in the experiments.

Tables~\ref{tab:annotated_person} and \ref{tab:trainset_annotation} provide statistics on the automatic labelling, showing complexity and diversity of the in-the-wild datasets.
In Table~\ref{tab:annotated_person}, one can find the maximum number of annotated bodies and faces on each corpus.
The reader can note that there is a mismatch between the number of detected bodies and faces across all datasets.
This is due to the performance of the underlying detection models that were trained on different data.
As expected, GAF and VGAF as GER datasets show the highest number of detected elements, and IER datasets may have more than 1 body or face detected.
Table~\ref{tab:trainset_annotation} gives insight about remaining training data after automatic annotation.
Indeed, videos for which no structural representation detection was performed are excluded from the training set.
Training data loss remained low (less than $3\%$ across all datasets). The automatic labelling process is not applied to the validation set.

\begin{table}%
\footnotesize\centering
\captionof{table}{Maximum number of bodies and faces annotated per dataset. No body annotations were done on DFEW.}
\label{tab:annotated_person}
\begin{autotabular}{l|cc}
\toprule
\multirow{2}[0]{*}{\textbf{Dataset}} & \multicolumn{2}{c}{\textbf{Maximum per frame}} \\
      & \textbf{Bodies} & \textbf{Faces} \\
\midrule
GAF-3.0 & 56    & 104 \\
VGAF  & 29    & \hphantom{0}71 \\
SAMSEMO & 18    & \hphantom{0}14 \\
MER2023 & 11    & \hphantom{0}16 \\
DFEW  & \mycross{} & \hphantom{00}1 \\
EngageNet & \hphantom{0}5     & \hphantom{00}1 \\
\bottomrule
\end{autotabular}
\vspace{0.5em}
\captionof{table}{Number of remaining videos in train/validation set after structural representation automatic annotation.}
\label{tab:trainset_annotation}
\begin{autotabular}{l|c|cc}
\toprule
{\multirow{2}[0]{*}{\textbf{Dataset}}} & \multirow{2}[0]{*}{\textbf{Validation-set}} & \multicolumn{2}{c}{\textbf{Train-set }} \\
      &       & \textbf{Original} & {\textbf{Automatic Annotation}} \\
\midrule
GAF-3.0 & 4356  & 9815  & 9558 (-2.62\%) \\
VGAF  & \hphantom{0}766   & 2661  & 2636 (-0.94\%) \\
SAMSEMO & \hphantom{0}412   & 9822  & 9762 (-0.61\%) \\
MER2023 & 2341  & 3373  & 3272 (-2.99\%) \\
DFEW  & 2098  & 9356  & 9353 (-0.03\%) \\
EngageNet & 1071  & 7879  & 7817 (-0.79\%) \\
\bottomrule
\end{autotabular}
\end{table}

\subsection{Preliminary experiments on Variational Encoder}

The preliminary experiments aim at designing two fundamental elements of VE-MD: the training strategy for the latent space \textit{Z} and the backbone architecture for the multi-task decoder. The following sections present these experiments.

\subsubsection{Training strategy}
\label{sec:residual_block}
The final definition of the encoder optimization resulted from a series of experiments conducted to identify the most suitable training strategy and the associated constraints, and therefore the corresponding loss functions, for training VE-MD. Three variants were investigated: a Variational Autoencoder (VAE), a vanilla encoder, and a Variational Encoder (VE). The VAE was trained using the ELBO objective, which combines a reconstruction term with KL-based latent regularization. In contrast, the vanilla encoder was trained without reconstruction and without any explicit latent-space constraint. The VE also removed the reconstruction objective, but retained latent-space regularization by matching the latent distribution to a Gaussian prior using an MMD loss~\cite{dziugaite2015training}. This choice was motivated by empirical observations: when the reconstruction term was removed, KL-only regularization resulted in unstable optimization, whereas MMD provided more stable training.

Experiments were conducted using a custom residual encoder (\textit{Custom}) composed of stacked residual downsampling blocks.
Given an input feature, the main branch of the residual block applies a stride-2
convolution followed by Batch Normalization (BN) and ELU activation, then a second convolution (stride 1) and BN.
In parallel, the skip connection branch uses a stride-2 convolutional projection to match both spatial resolution and channel dimension.
The block output is obtained by summation and a final ELU activation.

 \begin{table}[H]
 \centering
\caption{Encoder performance\protect\footnotemark~(VAE, VE, and Vanilla models) in terms of accuracy~(\%). Columns are sorted by mean accuracy on all datasets.}
\label{tab:vae_ve_result}
\begin{autotabular}{lccc}
  \toprule
  \textbf{Dataset}  & \textbf{VAE} & \textbf{Vanilla} & \textbf{VE} \\
  \midrule
  VGAF       & 48.05 & 48.96               & \textbf{51.54} \\
  MER-MULTI  & 32.52 & \textbf{40.29}      & 39.81 \\
  DFEW       & 44.62 & 53.92               & \textbf{56.22} \\
  SamSemo    & 57.19 & 57.43               & \textbf{62.68} \\
  EngageNET  & 54.20 & 55.32               & \textbf{56.12} \\
  \midrule
  Mean  & 47.32 & 51.58 & \textbf{53.67} \\
  \bottomrule
\end{autotabular}
 \end{table}
 \footnotetext{~\textbf{Bold values} indicate best score on validation set for each dataset.\label{fn:bold}}

Comparative results across five datasets (VGAF, DFEW, MER-MULTI, SAMSEMO, and EngageNet) are reported in Table~\ref{tab:vae_ve_result}. Overall, the VE configuration outperformed both the VAE and Vanilla variants on all datasets except MER-MULTI, where the performance difference remained marginal (0.28\%). The VE achieved performance gains of up to 6\% across datasets. In contrast, the Vanilla encoder, which does not enforce a Gaussian latent distribution, yielded weaker results, while the VAE showed lower emotion classification performance, likely because the reconstruction objective constrained the representation toward input reconstruction rather than discriminative emotion-related features. These findings suggest that enforcing a variational latent distribution while removing reconstruction constraints leads to a more compact and discriminative representation for emotion recognition, making the VE formulation particularly suitable for the multitask architecture adopted in VE-MD. Therefore, in the remaining experiments, the Variational Encoder was selected as the training strategy for our model.

\subsubsection{Multi-task encoder backbone selection}
\label{sec:vit_influence}

Beyond the custom residual encoder used in the previous VE experiments, this research investigates several publicly available pretrained CNN backbones, namely ResNet-50, ResNet-101, and ResNet-152, as a multi-task encoder alongside the ViT one.
The objective is to determine which architecture could provide more complementary representations to the ViT features.
In this experiment, the structural decoders are not used; the model is then a {Variational Encoder Single Decoder (VE-SD)}.

Table~\ref{tab:vit_ve_results} reports the results of VE-SD with ViT only and four variants combining it with \textit{Custom}, \textit{ResNet-50}, \textit{ResNet-101}, and \textit{ResNet-152}.
Overall, combining ViT with a residual encoder consistently improves performance over ViT only solution, confirming that the additional residual encoder provides useful complementary information.
Moreover, as expected, the ResNet-based variants outperform the \textit{Custom} variant for all datasets.
Among the ResNet variants, the differences are relatively small, with all three showing competitive performance across datasets.
\textit{ResNet-50} backbone achieves the best overall mean accuracy (71.01\%), slightly outperforming \textit{ResNet-101} (70.98\%) and \textit{ResNet-152} (70.62\%).
Although \textit{ResNet-101} obtains the best results on some individual datasets, notably DFEW and VGAF, \textit{ResNet-50} offers the best trade-off between accuracy and model complexity.
For this reason, \textit{ResNet-50} is selected as the multi-task encoder for all remaining experiments.

\begin{table}[t]
\centering
\caption{Accuracy\footref{fn:bold} (\%) results for ViT-only vs Variational Encoder combining ViT with different residual encoders. Columns are sorted by mean accuracy on all datasets.}
\label{tab:vit_ve_results}
\resizebox{\linewidth}{!}{%
\begin{autotabular}{l|c|cccc}
\toprule
\multirow{2}[0]{*}{\textbf{Dataset}} & \multirow{2}[0]{*}{\parbox{0.15\linewidth}{\centering\textbf{VE-SD}\\\textbf{ViT only}}} & \multicolumn{4}{c}{\textbf{VE-SD with ViT and Residual backbone}} \\
      &       & \textit{Custom} & \textit{ResNet-152} & \textit{ResNet-101} & \textit{ResNet-50} \\
\midrule
{GAF-3.0} & 82.27 & 82.41 & 82.41 & 82.33 & \textbf{83.56} \\
{VGAF} & 76.11 & 75.72 & 76.63 & \textbf{77.02} & 76.63 \\
{MER-MULTI} & 54.74 & 59.22 & 59.47 & \textbf{61.11} & 60.83 \\
{DFEW} & 64.88 & 68.53 & 68.64 & \textbf{70.02} & 64.69 \\
{SAMSEMO} & 55.38 & 66.02 & 69.62 & 67.90 & \textbf{72.26} \\
{EngageNet} & 64.14 & 67.82 & 66.92 & 67.35 & \textbf{68.10} \\
\midrule
Mean  & 66.25 & 69.95 & 70.62 & 70.98 & \textbf{71.01} \\
\bottomrule
\end{autotabular}%
}
\end{table}

\subsection{Experimental Setup and Training}
\label{sec:training_exp_setup}

{Variational Encoder Multi-Decoder (VE-MD)} receives as input a sequence of video frames.
All frames are resized to $224 \times 224$.
Frame sampling followed the temporal settings of each dataset.
For VGAF, 5 frames (1 per second) were kept as 5 frames per video because they provide the best accuracy in former experiments~\cite{augusma2023multimodal}.
In EngageNet, SAMSEMO, and  MER-MULTI, videos are longer (i.e varying from 1 to 41 seconds), thus 10 frames are uniformly selected to represent the video content.
Last, DFEW provides 16 frames for each video sequence.
To be able to compare with the state-of-the-art, the model uses an identical number of images as input.

For the multitask encoder branch, the \textit{ResNet-50} architecture produces a final feature map of size $4096 \times 7 \times 7$, which is reduced to $512 \times 7 \times 7$ ($C_z = 512$) using a $2$D convolution layer.
The same dimensionality reduction is applied to the ViT encoder to ensure consistent feature size.
Consequently, the final latent-space has dimensions $1024 \times 7 \times 7$. During training, the ViT encoder is first trained by finetuning on the target dataset using cross-entropy loss.
Then, during training of the multitask encoder and the structural decoders, it remains frozen.

The structural representation decoder uses two different sets of parameters.
When using the \textit{PersonQuery} decoder, three transformer layers are used, each with eight attention heads, and employ sinusoidal positional encoding for sequence representation.
The number of queries $Q_i$ determines the number of persons the model can detect, with $Q_i \in \{Q_{\text{Max}}, 50, 100\}$, where $Q_{\text{Max}}$ corresponds to the maximum number of annotated persons in the training dataset.
The loss function weights are set as follows: $\beta_{\text{limb}} = 1.0$ and $\beta_{\text{adj}} = 0.5$, giving higher importance to limb detection over adjacency refinement. Since emotion classification is the primary objective, the structural loss weights are kept lower, with $\beta_{p_i} = 0.1$ and $\beta_{\text{mmd}} = 0.1$ (see Equation~\ref{eq:total_loss}). The Adam optimizer is used with a learning rate of $10^{-7}$. %
    
For the \textit{Heatmap} decoder, the limbs decoder module is applied with six stages, each consisting of five CNN layers ($input=output=256$, $kernel=3$, $padding=1$).
The final outputs are heatmaps of size $output\_limbs \times 56 \times 56$, where $output\_limbs = 18$ for body connections and $83$ for facial landmarks.
It is worth noting that the \textit{VE-MD} with \textit{Heatmap} augments the original 63 connections with an additional 20 custom connections, resulting in a total of 83 connections.
The reason is that, unlike in the \textit{PersonQuery} approach, the heatmap dimension remains the same for the emotion decoder; it does not increase the model size.
For the loss function, weights are $\beta_{mmd}=0.1$ and $\beta_{p_i}=1.0$ (see Equation~\ref{eq:total_loss}), giving equal importance to structural representation detection and emotion classification.

\subsection{VE-MD including Structural Representation (VE-MD-SR)}
\label{sec:ve_md_sr_res}

In the VE-MD-SR setting, the predicted Structural Representation (SR) is provided as an additional input to the Emotion Decoder (ED), as illustrated by the orange arrow in Figure~\ref{fig:ve_md}. Two SR decoder variants are investigated: \textit{VE-MD-SR-PersonQuery} and \textit{VE-MD-SR-Heatmap}. For comparison, VE-MD variants trained in a multi-task manner but without injecting SR into the ED are analyzed later in the ablation study section (\autoref{sec:ablation_studies}).

Table~\ref{tab:VD_MD_SR_res} reports the best overall accuracy obtained by \textit{VE-MD-SR-PersonQuery} and \textit{VE-MD-SR-Heatmap} across the six GER and IER datasets, together with the single-decoder baseline \textit{VE-SD}. Both VE-MD-SR variants consistently outperform \textit{VE-SD}, confirming the benefit of explicitly incorporating structural representations into the emotion recognition pipeline. When comparing the two SR variants, neither consistently dominates across all datasets. \textit{VE-MD-SR-PersonQuery} achieves the best performance on GAF-3.0, MER-MULTI, SamSemo, and EngageNet, whereas \textit{VE-MD-SR-Heatmap} performs best on VGAF and DFEW. Importantly, the absolute performance gap ($|\Delta|$) between both variants remains small, ranging from 0.14 points on GAF-3.0 to 1.35 points on EngageNet, with an average below 1 point. The same overall trend is observed when considering GER and IER datasets separately, with each variant achieving the top result on half of the benchmarks.

This near-parity is particularly noteworthy because \textit{VE-MD-SR-Heatmap} offers two practical advantages over \textit{VE-MD-SR-PersonQuery}. First, it does not require fixing the number of persons at training or inference time, unlike the query-based formulation defined by $Q_{Body}$ and $Q_{Face}$. Second, it uses a constant parameterization, avoiding the additional learnable query embeddings required by the PersonQuery decoder. These properties make \textit{VE-MD-SR-Heatmap} a more flexible and lightweight alternative, especially in in-the-wild scenarios where the number of visible individuals is unknown or varies across video frames.

\begin{table*}%
    \centering
\caption{Overall accuracy\footref{fn:bold} (\%) of \textit{VE-MD-SR-PersonQuery} and \textit{VE-MD-SR-Heatmap} for all GER/IER datasets. Results are reported for best \textit{SR} configurations (Number of queries, use of STGCN, Body, Face, or Body + Face structural representation training) for each dataset. $Q$ Body and $Q$ Face denote the number of respective queries, with $Q_{Max}$ corresponding to the maximum number of annotated bodies/faces in the dataset (\qmaxval{underlined values}). $\boldsymbol{|\Delta|}$ represents absolute difference between \textit{PersonQuery} and \textit{Heatmap} best performances.}
    \label{tab:VD_MD_SR_res}
    \begin{autotabular}{l|c|ccccc|cc|c}
    \toprule
    {\multirow{2}[2]{*}{\textbf{Dataset}}} & \multirow{2}[2]{*}{\textbf{VE-SD}} & \multicolumn{5}{c}{\textbf{\textit{VE-MD-SR-PersonQuery}}} & \multicolumn{2}{c}{\textbf{\textit{VE-MD-SR-Heatmap}}} & \multirow{2}[2]{*}{\textbf{$\boldsymbol{|\Delta|}$}} \\
          &       & \textbf{SR training} & \textbf{use STGCN} & \textbf{$\boldsymbol{Q}$ Body} & \textbf{$\boldsymbol{Q}$ Face} & \textbf{Accuracy} & \textbf{SR training} & \textbf{Accuracy} &  \\
    \midrule
    GAF-3.0 & 83.56 & Body + Face & yes   & 100   & 100   & \textbf{90.06 ($\uparrow$6.50)} & Body + Face & 89.92 ($\uparrow$6.36) & 0.14 \\
    VGAF  & 76.63 & Body  & no    & 50    & \mycross{}     & 78.46 ($\uparrow$1.83) & Face  & \textbf{79.77 ($\uparrow$3.14)} & 1.31 \\
    \midrule
    MER-MULTI  & 60.83 & Body + Face  & yes    & 16    & 16     & \textbf{62.38 ($\uparrow$1.55)} & Body + Face & 61.89 ($\uparrow$1.06) & 0.49 \\
    DFEW  & 64.69 & Face  & no    & \mycross{}     & \qmaxval{1} & 68.96 ($\uparrow$4.27) & Face  & \textbf{70.31 ($\uparrow$5.62)} & 1.3 \\
    SamSemo & 72.26 & Body  & no   & \qmaxval{18} & \mycross{}     &\textbf{ 75.29 ($\uparrow$3.03)} & Body + Face & {75.10 ($\uparrow$2.84)} & 0.19 \\
    EngageNet & {68.10} & Body + Face  & yes    & \qmaxval{1}     & \qmaxval{1} & \textbf{68.98} ($\uparrow$0.88) & Body  & 67.63 ($\downarrow$0.47) & 1.35 \\
    \bottomrule
    \end{autotabular}%

\end{table*}

\section{Ablation studies}
\label{sec:ablation_studies}

\subsection{Variational Encoder with Multi Decoders only (VE-MD)} 
\label{sec:ve_md_sr_ablation}

Before investigating \textit{VE-MD-SR}, in which structural representation (SR) is also provided as input to the emotion decoder, we first study the \textit{VE-MD} setting. In this configuration, all variants are trained in a multi-task manner: the structural decoders are used to guide and optimize the $z_2$ embedding, but their SR outputs are not fed to the emotion decoder during training. In other words, the SR-to-emotion-decoder connection illustrated by the orange arrow in Figure~\ref{fig:ve_md} is removed in this ablation. At inference time, the structural decoders are also discarded. This setting further supports the privacy-aware aspect of the proposed method, since no structural information is decoded at inference time. As in the previous experiments, two variants are evaluated: \textit{VE-MD-PersonQuery} and \textit{VE-MD-Heatmap}.

Table~\ref{tab:VD_MD_res} reports the best results obtained by the \textit{VE-MD} variants. Across all six datasets, \textit{PersonQuery} consistently matches or outperforms \textit{Heatmap}, achieving the best accuracy in five out of six cases. At the same time, the absolute difference $|\Delta|$ between both variants remains small across datasets (0.35--1.68), showing that \textit{Heatmap} remains a competitive and more flexible alternative, while avoiding the need to specify a query count.

Despite the absence of SR input to the emotion decoder, both variants still improve over \textit{VE-SD} on all IER datasets, most notably on DFEW (+5.29 for \textit{PersonQuery}) and SamSemo (+3.38), indicating that the multi-person design itself provides a meaningful benefit for emotion recognition. By contrast, the gains are limited on the GER datasets: only a small improvement is observed on VGAF with \textit{PersonQuery}, while performance on GAF-3.0 falls below the \textit{VE-SD} baseline. Compared with the SR-equipped configurations reported in Table~\ref{tab:VD_MD_SR_res}, this ablation suggests that incorporating SR input into the emotion decoder provides additional benefits, particularly for GER datasets.

\begin{table*}[t]
  \centering
  \caption{Overall accuracy\footref{fn:bold} (\%) of \textit{VE-MD-PersonQuery} and \textit{VE-MD-Heatmap} for all GER/IER datasets. Results are reported for the best SR training configurations (Number of queries, Body, Face, or Body + Face structural representation training) for each dataset. $Q$ Body and $Q$ Face denote the number of respective queries, with $Q_{Max}$ corresponding to the maximum number of annotated persons/faces in the dataset (\qmaxval{underlined values}). $\boldsymbol{|\Delta|}$ represents absolute difference between \textit{PersonQuery} and \textit{Heatmap} best performances.}
    \begin{autotabular}{l|c|Hcccc|cc|c}
    \toprule
    {\multirow{2}[2]{*}{\textbf{Dataset}}} & \multirow{2}[2]{*}{\textbf{VE-SD}} & \multirow{2}[2]{*}{\makecell{\textbf{Best VE-MD}\\\textbf{SR variant}}} & \multicolumn{4}{c}{\textbf{VE-MD-PersonQuery}} & \multicolumn{2}{c}{\textbf{VE-MD-Heatmap}} & \multirow{2}[2]{*}{\boldmath{}\textbf{$|\Delta|$}\unboldmath{}} \\
          &       &       & \textbf{SR training}  & \textbf{Q Body} & \textbf{Q Face} & \textbf{Accuracy} & \textbf{SR training} & \textbf{Accuracy} &  \\
    \midrule
    GAF-3.0 & \textbf{83.56} & 90.06 & Body      & 50    & \mycross{}     & 82.37 ($\downarrow$1.19) & Face  & 82.95 ($\downarrow$0.61) & 0.53 \\
    VGAF  & 76.63 & 79.77 & Body      & 50    & \mycross{}     & \textbf{77.28 ($\uparrow$0.65)} & Body  & 76.37 ($\downarrow$0.26) & 0.91 \\
    \midrule
    MER-MULTI  & 60.83 & 62.14 & Body + Face     & \qmaxval{16}   &\qmaxval{16}      & \textbf{62.38 ($\uparrow$1.55)} & Face  & 61.65 ($\uparrow$0.82) & 0.73 \\
    DFEW  & 64.69 & 70.31 & Face      & \mycross{}     & \qmaxval{1}   & \textbf{69.98 ($\uparrow$5.29)} & Face  & 68.30 ($\uparrow$3.61) & 1.68 \\
    SamSemo & 72.26 & 75.10 & Body      & 50    & \mycross{}     & \textbf{75.64 ($\uparrow$3.38)} & Face  & 75.29 ($\uparrow$3.03) & 0.35 \\
    EngageNet & 68.10 & 68.98 & Face      & \mycross{}     & \qmaxval{1}   & \textbf{68.47 ($\uparrow$0.37)} & Body + Face & 68.10 (=0.00) & 0.37 \\
    \bottomrule
    \end{autotabular}%
  \label{tab:VD_MD_res}%
\end{table*}%

\subsection{Ablations on Structural Representation}
In the proposed approach, structural representation (SR) can affect performance through several design choices. First, SR can either be used only as an auxiliary supervision signal to optimize the latent space ($z_2$) in the multi-task setting, or be additionally provided as input to the emotion decoder, as explored in Sections~\ref{sec:ve_md_sr_res} and~\ref{sec:ve_md_sr_ablation}. 
When SR is used as input to the emotion decoder, it can also be projected into a lower-dimensional representation, which may further influence performance. 
Second, in the \textit{PersonQuery} variant, performance may depend on the number of queries defined at training time and on whether STGCN is used. 
Third, the type of structural representation itself may play a role, depending on whether body, face, or body+face information is employed. 
This subsection presents ablation studies to assess the impact of these SR-related design choices.

\subsubsection{Effect of SR Projection}
\label{sec:projection_effect}

We investigate whether using the raw SR output or a projected SR representation is more effective. The results reveal opposite behaviors for GER and IER datasets. To illustrate this difference and identify a suitable projection dimension, we first consider two representative datasets: VGAF for GER and SamSemo for IER. In these experiments, body-based SR is used. For this analysis, we select the \textit{VE-MD-SR-Heatmap} decoder because it is more convenient than \textit{VE-MD-SR-PersonQuery}, which requires fixing the query count. The raw SR outputs are linearly projected to match the latent-space dimension. Without projection, the SR embedding size is $2 \times latent\_dim + 56 \times 56$. With projection, a multiplicative factor in $\{0.5,1,2,3,4\}$ is used to scale the projection dimension, resulting in an embedding size of $2 \times latent\_dim + \text{factor} \times latent\_dim$.

The results are reported in Table~\ref{tab:proj_heatmap_test_dim}. Across the tested factors, the best projection size is 512, corresponding to $\text{factor}=1$. On VGAF (GER), the best accuracy is obtained without projection (79.77\%), and performance decreases when projection is introduced. By contrast, on SamSemo (IER), the accuracy without projection is 67.48\%, and performance improves steadily with projection, reaching 74.81\%. After identifying the best projection size, we extend the analysis to face-based SR and include additional datasets, namely MER-MULTI and GAF-3.0, in order to confirm the trend observed between GER and IER datasets. The corresponding results are reported in Table~\ref{tab:proj_heatmap_results}. For GER, raw SR consistently outperforms projected SR (+7\% on GAF-3.0 and +2\% on VGAF), whereas for IER, projection acts as a beneficial bottleneck, yielding gains of about +8\% on MER-MULTI and +10\% on SamSemo.

\begin{table}[H]
\centering
\caption{Accuracy (\%) comparison for projection-size (Proj.) with \textit{VE-MD-SR-Heatmap} decoder for body SR. Evaluation for VGAF (GER) and SAMSEMO (IER).}
\label{tab:proj_heatmap_test_dim}
\resizebox{\linewidth}{!}{
    \begin{autotabular}{ccccccc}
    \toprule
    \textbf{Dataset} & \textbf{latent\_dim} & \textbf{L\_Proj} & \textbf{Factor} & \textbf{Proj\_size} & \textbf{Emb\_size} & \textbf{Acc. (\%)} \\
    \midrule
    \multirow{6}[0]{*}{\begin{sideways}VGAF\end{sideways}} & 1024  & No    & \mycross{}     & \mycross{}     & 5184  & 78.20 \\
          & 512   & No    & \mycross{}     & \mycross{}     & 4160  & \textbf{79.7}7 \\
          & 512   & Yes   & 4     & 2048  & 3072  & 77.81 \\
          & 512   & Yes   & 3     & 1536  & 2560  & 78.07 \\
          & 512   & Yes   & 2     & 1024  & 2048  & 77.94 \\
          & 512   & Yes   & 1     & \phantom{0}512   & 1536  & 77.42 \\
    \midrule
    \multirow{6}[0]{*}{\begin{sideways}SAMSEMO\end{sideways}} & 512   & No    & \mycross{}     & \mycross{}     & 4160  & 67.48 \\
          & 512   & Yes   & 4     & 2048  & 3072  & 73.05 \\
          & 512   & Yes   & 3     & 1536  & 2560  & 72.81 \\
          & 512   & Yes   & 2     & 1024  & 2048  & 73.29 \\
          & 512   & Yes   & 1     & \phantom{0}512   & 1536  & \textbf{74.81} \\
          & 512   & Yes   & 0.5   & \phantom{0}256   & 1280  & 73.57 \\
    \bottomrule
    \end{autotabular}%
}
\end{table}

\begin{table}[H]
\centering
\caption{Accuracy (\%) comparison between projected and non-projected face SR in \textit{VE-MD-SR-Heatmap} on GER and IER datasets.}
\label{tab:proj_heatmap_results}
\begin{autotabular}{l|cc}
\toprule
\textbf{Dataset} & \textbf{Acc. (Projected)} & \textbf{Acc. (Without projection)} \\ 
\midrule
GAF-3.0   & 82.56 & \textbf{89.60} \\
VGAF      & 77.42 & \textbf{79.77} \\
\midrule
MER-MULTI & \textbf{61.17} & 54.85 \\
SamSemo   & \textbf{75.00} & 55.34 \\
\bottomrule
\end{autotabular}
\end{table}

\subsubsection{Structural Representation Modality}

As described previously, three structural representation (SR) modalities can be considered in the proposed \textit{VE-MD} architecture: body, face, and body+face. This ablation evaluates the impact of these modalities on both \textit{VE-MD-SR-PersonQuery} and \textit{VE-MD-SR-Heatmap}. Table~\ref{tab:resnet_detr_SR} reports the influence of SR modality across the GER/IER datasets for \textit{VE-MD-SR-PersonQuery}. The results show that the best modality is dataset-dependent, suggesting that the relative importance of body and face cues varies according to the visual content and interaction characteristics of each benchmark.

On GAF-3.0, the highest performance is obtained with Body+Face combined with STGCN (90.06), clearly outperforming the body-only and face-only settings. This indicates a strong complementarity between both modalities on this dataset. By contrast, VGAF achieves its best result with Body only at $Q_{50}$ without STGCN (78.46), showing that adding facial information or graph refinement is not systematically beneficial. A similar dataset-dependent behavior is observed on the IER datasets. On MER-MULTI, the highest accuracy (62.38) is obtained by both the Body and Body+Face configurations when combined with STGCN.
DFEW strongly favors face cues, with the best accuracy obtained using Face-only SR without STGCN (68.96). SamSemo reaches its best result with Body at $Q_{\max}$ (75.29), whereas EngageNet performs best with Body+Face combined with STGCN (68.98). Overall, these results indicate that no single SR modality consistently dominates across datasets. Instead, the most effective modality depends on the characteristics of the benchmark, which makes the selection of SR modality dataset-specific. Interestingly, this is not always fully predictable from the apparent visual framing of the data. For instance, although EngageNet mainly contains head-centered videoconferencing views, the best result is still obtained with Body+Face.

Modality ablation results for \textit{VE-MD-SR-Heatmap} are reported in Table~\ref{tab:SR_type_heatmap}. Similar trends are observed. The three modalities do not improve performance uniformly across datasets. For the GER datasets, the combination of Body+Face improves performance on GAF-3.0, but not on VGAF, where Face-only SR yields the best result. One possible explanation is that GAF-3.0 consists of static images, where body and face cues can be jointly exploited more consistently, whereas in video datasets, facial visibility may vary across frames. For the IER datasets, the results suggest a more complementary contribution of both modalities, as observed for MER-MULTI and SamSemo, where Body+Face achieves the best performance. In DFEW, only the face modality is available, as shown in Figure~\ref{fig:datasets_used_overview}, which explains why this dataset is evaluated only with Face-based SR.

\begin{table*}%
    \centering
\caption{Accuracy (\%) ablations on \textit{VE-MD-SR-PersonQuery} for GER datasets. Results are reported for different SR configurations (Body, Face, and Body+Face) across query settings. $Q_{i}$ denotes the number of queries, with $Q_{Max}$ corresponding to the maximum number of annotated persons in the dataset, $Q_{50}$: $query=50$,  $Q_{100}$:~$query=~100$. The best query value is selected to train body+face. }
    \label{tab:resnet_detr_SR}
    \renewcommand{\arraystretch}{1.4} %
    
    \begin{autotabular}{@{}l|ccc|ccc|ccc|ccc|cc@{}}
        \toprule
         \textbf{Datasets} & \multicolumn{6}{c|}{\textbf{Body}} & \multicolumn{6}{c|}{\textbf{Face}} & \multicolumn{2}{c}{\textbf{Body+Face}} \\ 
        \cmidrule{2-15}

      & \multicolumn{3}{c|}{  SR} & \multicolumn{3}{c|}{  SR+STGCN} &  \multicolumn{3}{c|}{  SR} & \multicolumn{3}{c|}{  SR+STGCN}  &  \multicolumn{1}{c|}{  SR} & \multicolumn{1}{c}{  SR+STGCN}  \\ 

        \cmidrule{2-15}
         &  $Q_{Max}$ &   $Q_{50}  $ &   $Q_{100}$ &   $Q_{Max}$ &   $Q_{50}  $ &   $Q_{100}$ &   $Q_{Max}$ &   $Q_{50}  $ &   $Q_{100}$ &   $Q_{Max}$ &   $Q_{50}  $ &   $Q_{100}$   &   & \\ 
        
        \midrule
        GAF-3.0   & {84.38} & {84.22} & {84.89}   &{86.45}&{86.14}&{87.99}  &{85.72}&{85.40}&{86.11}  &{87.67}&{86.27}&{88.25} &{85.99}&\textbf{90.06} \\ 
        VGAF      &{75.98}&\textbf{78.46}&{{77.55}}  &{77.02}&{77.02}&{78.33}  &{77.28}&{77.28}&{77.55} &{77.15}&{76.11}&{76.5}  &{78.33} &{77.55} \\ 
        \midrule
        MER-MULTI    &{61.17} &{61.41} & \mycross{} &\textbf{62.38}&{60.19}   & \mycross{}&{60.19}&{61.41}& \mycross{}    &{62.14}&{61.65}  & \mycross{}&{60.92} &\textbf{62.38} \\ 
        DFEW  & \mycross{}& \mycross{}& \mycross{}& \mycross{}& \mycross{} & \mycross{} &\textbf{68.96}&{64.45}   & \mycross{} &{68.66}& \mycross{}& \mycross{}& \mycross{}& \mycross{}\\ 
        SamSemo  &\textbf{75.29}&{74.76} & \mycross{}&{74.29} & \mycross{}& \mycross{}&{75.14} &{75.19} & \mycross{}  &{74.38}&{74.63}  & \mycross{}&{75.05} &{74.29} \\ 
       EngageNet   &{{68.53}}&{67.35} & \mycross{}  &{67.82}&{67.98} & \mycross{} &{67.91}&{67.63} & \mycross{}&{68.19}&{66.88} & \mycross{} &{68.38} &\textbf{68.98} \\ 
        \bottomrule
        
    \end{autotabular}
\end{table*}

\begin{table}%
  \centering
    \caption{Accuracy (\%) for the ablation study on the influence of structural representation type in \textit{VE-MD-SR-Heatmap}.}%
    \begin{autotabular}{l|ccc}
    \toprule
    \textbf{Dataset} &\textbf{Body} & \textbf{Face} & \textbf{Body + Face} \\
    \midrule
    GAF-3.0  & {86.72} & 89.60 & \textbf{89.92} \\
    VGAF   & {79.24} & \textbf{79.77} & 79.37 \\
    \midrule
    MER-MULTI  & 61.17 & 60.19 & \textbf{61.89} \\
    DFEW   & \mycross{} & \textbf{70.31} & \mycross{} \\
    SamSemo  & 74.81 & 75.00    & \textbf{75.10} \\
    EngageNet &\textbf{67.63} & 67.26 & 67.26 \\
    \bottomrule
    \end{autotabular}%
  \label{tab:SR_type_heatmap}%
\end{table}%

\subsubsection{Query Count and Use of STGCN}
For the \textit{PersonQuery} approach (see Table~\ref{tab:resnet_detr_SR}), the main observation is that performance is sensitive to the query count ($Q$). On GAF-3.0, performance generally improves as $Q$ increases, reaching its best value at $Q=100$. However, this trend does not hold for all datasets. On VGAF, for example, using $Q_{\max}$ ($Q=29$) yields 75.98\%, increasing the number of queries to $50$ improves performance to 78.46\%, whereas further increasing it to $100$ reduces performance to 77.55\% for the body-based structural representation. A similar pattern is observed for the Body+Face configuration. One possible explanation is that using a fixed number of queries during training does not always match the actual number of persons present in the videos. Moreover, even when the selected query count matches the number of persons in one frame, this correspondence may no longer hold in subsequent frames due to temporal variation in visibility and scene composition. 

A related trend is observed on the IER datasets, where most benchmarks achieve their best performance when using $Q_{Max}$. In most cases, increasing the query count to $50$ leads to a decrease in performance. For example, DFEW achieves 68.96 with $Q_{Max}$ but only 64.45 with $Q_{50}$, while MER-MULTI reaches 62.38 with $Q_{Max}$ and drops to 60.19 with $Q_{50}$. This behavior is consistent with the fact that the emotional content in these datasets is often centered on a single main person, making larger query counts less suitable. For this reason, experiments with $Q=100$ were not conducted on these datasets.

Regarding STGCN, the results show that its contribution is not consistent across either GER or IER datasets. It helps improve performance on GAF-3.0, where the best result reaches 90.06, but it does not provide the same benefit on VGAF. A similar pattern is observed on the IER datasets: STGCN contributes to the best performance on MER-MULTI (62.38) and EngageNet (68.98), but not on DFEW. One possible explanation is the noise and temporal inconsistency in the automatically generated annotations, since some persons may be inaccurately detected or not tracked consistently across frames. To address these limitations, the \textit{Heatmap} approach is introduced as an alternative that can better accommodate variations in the number of people across videos and across frames.

\subsection{Ablation on ViT Branch Pretraining}

For the GER datasets, we additionally investigate whether initializing the ViT branch with a model pretrained for group emotion recognition improves performance. To this end, we use the synthetic-video pretraining strategy introduced in our previous work~\cite{augusma2023multimodal}. In that setting, synthetic videos are generated by compositing face images with controlled expressions, selected from FACES and KDEF~\cite{Ebner2010,calvo2018human}, onto diverse LSUN backgrounds~\cite{Yu2015}. This process produces group emotion-labeled frames while preserving group-level affect, and was previously shown to encourage the model to focus on facial cues relevant to group emotion~\cite{augusma2023multimodal}.

We reuse this pretrained ViT branch to train \textit{VE-MD-SR-Heatmap} on the GER datasets. The results are reported in Table~\ref{tab:heatmap_skt_results}. Synthetic-video pretraining improves performance on VGAF, where the best accuracy increases from 79.77\% to 80.81\% (+1.04), but slightly decreases performance on GAF-3.0, where the best result drops from 89.92\% to 89.55\% (-0.37). Overall, these results suggest that synthetic-video pretraining can be beneficial, although its impact remains dataset-dependent. We include this initialization setting to enable direct comparison with our previous work~\cite{augusma2023multimodal}.

 \begin{table}[H]
     \centering
\caption{Best accuracy (\%) of \textit{VE-MD-SR-Heatmap} on GER datasets with and without synthetic-video pretraining of the ViT branch.} \label{tab:heatmap_skt_results}
     \begin{autotabular}{@{}l|ccc|ccc@{}}
         \toprule
         \textbf{Dataset} & \multicolumn{3}{c|}{\textbf{Pretrained ViT (no-Synt)}} & \multicolumn{3}{c}{\textbf{Pretrained ViT (with-Synt)}} \\
         \cmidrule(lr){2-7}
           & Body & Face  & Body+Face  & Body & Face  & Body+Face\\ 
         \midrule
         GAF-3.0 & 86.72 & 89.60 & \textbf{89.92} & 86.36&89.23&\textbf{89.55} \\
         VGAF & 79.24 & \textbf{79.77} & 79.37  & 80.68 & \textbf{80.81} & 80.55  \\

         \bottomrule
     \end{autotabular}
 \end{table}

\section{Discussion and Limitations}

\subsubsection{Discussion}

The ablation studies support the central hypothesis of this work: structural supervision improves emotion recognition, but its contribution depends on how structural information is represented and integrated, and on whether the target task is GER or IER.

Both decoder variants, \textit{PersonQuery} and \textit{Heatmap}, confirm the value of joint structural-affective learning. However, \textit{PersonQuery} remains constrained by its fixed-query formulation, since performance depends on the selected query count and does not always benefit from STGCN. These constraints are particularly problematic when the number of visible persons varies across videos and across frames. In contrast, \textit{Heatmap} naturally handles variable group size without requiring predefined queries, making it a more flexible and stable design.

A key result is that latent-space optimization alone is not sufficient for GER. When structural decoders are used only as auxiliary supervision in \textit{VE-MD}, the gains remain limited on GER datasets, although clear improvements are observed on IER benchmarks. By contrast, when SR outputs are explicitly provided to the emotion decoder in \textit{VE-MD-SR}, performance improves substantially on GER. This indicates that compressing structural information only into the shared latent representation tends to attenuate interaction-related cues that are important for collective affect modeling. In GER, emotion depends not only on isolated body or face evidence, but also on inter-person relations. Preserving explicit structural outputs up to the decoder, therefore, appears necessary to retain this relational information.

The modality ablations further show that no single SR modality consistently dominates across all datasets. The best choice among body, face, and body+face remains dataset-dependent, indicating that the contribution of structural cues varies with the visual and interaction characteristics of each benchmark. Similarly, the projection study reveals opposite behaviors for GER and IER. For GER, raw SR consistently outperforms projected SR, suggesting that projection removes useful relational information. For IER, projection instead acts as a beneficial bottleneck, improving performance by suppressing noisy or redundant structural cues. This distinction reinforces the idea that collective affect recognition requires richer intermediate relational representations, whereas individual emotion recognition benefits more from compact and denoised structural information.

The ViT pretraining ablation also points to a dataset-dependent effect. Synthetic-video pretraining improves performance on VGAF but not on GAF-3.0, suggesting that such initialization can be beneficial, but is not universally advantageous.

Last, findings on structural representation projection highlight a fundamental distinction between individual and collective affect modeling. For IER, compact latent representations and projected structural cues are often sufficient, since the emotional content is typically centered on a dominant individual. For GER, however, preserving explicit structural information is more important, as group emotion emerges from inter-person dynamics and scene-level configuration. Importantly, this is achieved while maintaining a privacy-aware design: the framework predicts only group-level emotion and does not require identity modeling or explicit per-person emotion inference as a final objective.

\subsubsection{Limitations}

While VE-MD demonstrates strong performance across GER and IER datasets, several limitations remain. First, the proposed privacy-aware design restricts the model to group-level predictions, but it does not provide formal anonymization of the input data or cryptographic privacy guarantees, since the system operates directly on full RGB frames. Second, the structural supervision depends on automatically generated pose and landmark annotations, which may introduce noise and bias into the training process. Third, although explicitly incorporating structural outputs improves GER performance, it also increases architectural complexity compared with approaches that rely only on latent representations. Future work will investigate interaction-aware latent disentanglement mechanisms able to encode relational cues directly within the shared representation, with the goal of reducing reliance on explicit structural outputs while preserving collective affect modeling performance.

\section{VE-MD as video encoder for multimodal emotion recognition}
\label{sec:multimodal}

\subsection{Combining VE-MD with Audio}
\label{sec:multimodal_ve_md}
To enable comparison with multimodal state-of-the-art methods on \textit{VGAF}, \textit{SAMSEMO}, and \textit{MER-MULTI}, we extend VE-MD with an audio branch. This extension is motivated by the fact that emotional content is not only conveyed visually, but also through lexical information and paralinguistic vocal cues.

\subsubsection{Audio Modeling}

We use two complementary audio encoders: (i) a \emph{content} encoder and (ii) an \emph{acoustic} representation encoder. Content encoders capture lexical and semantic information correlated with emotion, whereas acoustic encoders capture paralinguistic cues directly from the speech signal.

For content modeling, we use \emph{Whisper (medium)}~\cite{radford2023robust}. For acoustic representation learning, we use \emph{Wav2Vec~2.0}~\cite{baevski2020wav2vec}. Whisper is a transformer-based multilingual ASR model trained at scale, providing robust text-related audio representations under noisy conditions. Wav2Vec~2.0 learns contextualized speech features from raw waveforms in a self-supervised manner and has shown strong performance on speech and paralinguistic tasks. When both encoders are used jointly, we model their interaction through bidirectional cross-attention. Let $A$ denote the acoustic representation and $C$ the content representation. We compute: $\mathrm{Att}_{ACC} = \mathrm{Attention}(A, C, C),$ and $\mathrm{Att}_{CAA} = \mathrm{Attention}(C, A, A).$

The final audio representation is obtained by concatenating $A$, $C$, $\mathrm{Att}_{ACC}$, and $\mathrm{Att}_{CAA}$.

\subsubsection{Audio-Video Fusion Strategy}
To combine the audio and VE-MD branches, the final design uses \textit{late fusion}: each branch is first linearly projected to a common dimension, followed by multi-head self-attention and a FAP layer for sequence reduction. The fused representation is then fed to an MLP for classification. In addition to the late fusion, \textit{Attention-Guided Feature (AFG)}~\cite{wang2023hierarchical} is evaluated. The \textit{AFG} is a lightweight attention-based fusion module used to combine the audio and video embeddings into a single multimodal representation. Given an audio feature $f_i^{a}\in\mathbb{R}^{d_a}$ and a video feature $f_i^{v}\in\mathbb{R}^{d_v}$, AFG first projects both features into a shared $D$-dimensional space to obtain aligned vectors $h_i^{a}, h_i^{v}\in\mathbb{R}^{D}$. A small MLP with a softmax predicts two fusion weights $\alpha_i^{a}$ and $\alpha_i^{v}$ ($\alpha_i^{a}+\alpha_i^{v}=1$), which reflect the relative importance of audio vs.\ video for sample $i$. The final fused feature is computed as a weighted sum: $\hat{h}_i = \alpha_i^{a} h_i^{a} + \alpha_i^{v} h_i^{v}.$ This allows the model to rely more on the modality that is more informative (or less noisy) for each input.

\subsubsection{Training and Experimentation}

To train the audio branch, the corresponding backbone is kept frozen. A linear layer is added in the last sequence of the embedding to project audio features to 1024, then classified with an MLP. The Adam optimizer is used with the learning rate fixed to $1e-0.4$ using cross-entropy loss. For audio-video fusion, \textit{VE-MD} is also kept frozen before feeding the multimodal emotion classification. The best \textit{VE-MD} video for each dataset is selected in terms of performance. We observe that the video features that perform best in isolation do not always lead to the best fusion with a given audio branch, likely because the two modalities may capture overlapping rather than complementary information. For VGAF, the best audio-visual combination is obtained with the final \textit{VE-MD-SR-Heatmap} version using Body+Face. For MER-MULTI, the best fusion result is achieved by \textit{VE-MD-PersonQuery} with Body+Face, whereas for SamSemo, the best configuration uses Body only.

The summary of best performance for audio-video fusion is given in Table~\ref{tab:multimodal_audio_video}. On the VGAF dataset, the best performance is 82.25\% of accuracy when combining Wav2Vec~2.0 and video via late fusion, combining with the AFG module. For MER-MULTI, the best configuration combines Whisper+Wav2Vec~2.0 + video branch, giving 64.23\% of accuracy. In the case of SAMSEMO, the video combination with Whisper achieves 78.07\% of accuracy.

\begin{table}[H]
  \footnotesize
  \centering
  \caption{Summary of Audio-Video Fusion Accuracy (\%).}
  \label{tab:multimodal_audio_video}
  \begin{autotabular}{@{}l|cccc@{}} %
  \toprule
  \textbf{Datasets} & \textbf{Audio-Encoder} & \textbf{Acc. Audio} & \textbf{Acc. Video} & \textbf{Acc. Fusion} \\
  \midrule
  VGAF &  Wav2Vec~2.0 & 58.49 & 80.55 & \textbf{82.25} \\
  
   MER-MULTI & {Wav2Vec~2.0 + Whisper} &
  63.75 & 62.38 & \textbf{64.23} \\

SAMSEMO & {Wav2Vec~2.0 + Whisper} & 74.78 & 75.64 & \textbf{78.07} \\
  \bottomrule
  \end{autotabular}
\end{table}

\subsection{Combining VE-MD with Audio and text}
For SAMSEMO, in order to compare with the SOTA, we train the model with a text branch extension. For text features, we use Mixtral-8x7B and Llama-2-7B.  %
Both produce 4096-dimensional embeddings, which are projected to 1024 for alignment. The performances for the text branch are 60.18\% and 62.69\% of accuracy for Llama-2 and Mixtral, respectively. When adding text to the audio–video fusion, performance improves further. The best result achieved with Mixtral features is 79.18\% accuracy for 77.94\% of weighted F1. The best performance with Llama-2 features is 79.20\% of accuracy for 77.71\%  of F1-score.

Finally, a summary of the multimodal extension combining \textit{VE-MD} and \textit{audio} (or \textit{text}) branches is given in the Table~\ref{tab:acc_f1_summary} with the best performance accuracy and corresponding F1-score. As an observation, a well-balanced distribution across classes is shown for GAF-3.0, VGAF, and MER-MULTI based on the F1-scores, which is different from other datasets.

\begin{table}[H]
\centering
\caption{Multimodal performance's summary per dataset. Accuracy and corresponding  F1-score, (in \%).}
\label{tab:acc_f1_summary}
\begin{autotabular}{lcc}
\toprule
\textbf{Dataset} & \textbf{Accuracy} & \textbf{F1-score} \\
\midrule
GAF3.0     & 90.06 & 89.92 \\
VGAF       & 82.25 & 82.16 \\
\midrule
SamSemo    & 79.20 & 77.94 \\
MER-MULTI  & 64.94 & 64.84 \\
DFEW       & 70.73 & 68.20 \\
EngageNet  & 68.98 & 67.54 \\
\bottomrule
\end{autotabular}
\end{table}

\section{Comparisons with State-of-the-Art}
\label{sec:comparison_sota}
Having validated the \textit{VE-MD} framework across visual and multimodal configurations, we now benchmark its performance against existing state-of-the-art  (SOTA) methods and baselines across all datasets.
Comprehensive comparison and ranking against SOTA systems on all datasets are detailed in section \ref{apdx:sota} ``\nameref{apdx:sota}''.

\subsection{SOTA Comparison for GER Datasets}

In Figure~\ref{fig:sota_graph}, it is shown for GAF-3.0 that \textit{VE-MD} surpasses significantly prior work by a clear margin. We report accuracy along with 95\% confidence intervals computed using the Wilson score interval. 
For pairwise comparisons between methods evaluated on the same test set, statistical significance is assessed using a two-sided McNemar's test, with $p < 0.05$ considered significant. Our best configuration (body and face structures) improves accuracy by {+3.16 points} over the last prior work (90.06\% vs.\ 86.90\%). Even using a single predicted structural representation decoder (face \emph{or} body), VE-MD exceeds SOTA~\cite{wang2018cascade}, underscoring its effectiveness for group emotion recognition in images.

It is worth noting that Zhu et al.~(2025)~\cite{zhu2025adaptive} recently proposed the Key Role Guided Hierarchical Relation Inference (KR-HRI) model to enhance group-level emotion recognition. This method is not included in the comparison presented in Figure~\ref{fig:sota_graph}, as it reports performance using a different evaluation metric, Unweighted Average Recall (UAR), rather than accuracy. On the GAF-3.0 dataset, KR-HRI achieved a UAR of 81.29\%. In contrast, our best \textit{VE-MD} model reached a UAR of {88.70\%} on GAF-3.0, outperforming KR-HRI by {+7.41} percentage points.

For the VGAF dataset, \textit{VE-MD} also proves effective. Our multimodal approach (VE-MD + Wav2Vec~2.0) achieves {82.25\%} accuracy, improving over the last prior work (81.98\%)~\cite{kumar2020object} by {+0.27 points}.

\subsection{SOTA Comparison for IER Datasets} 
\begin{figure}[H]
  \centering
    \includegraphics[width=0.95\linewidth]{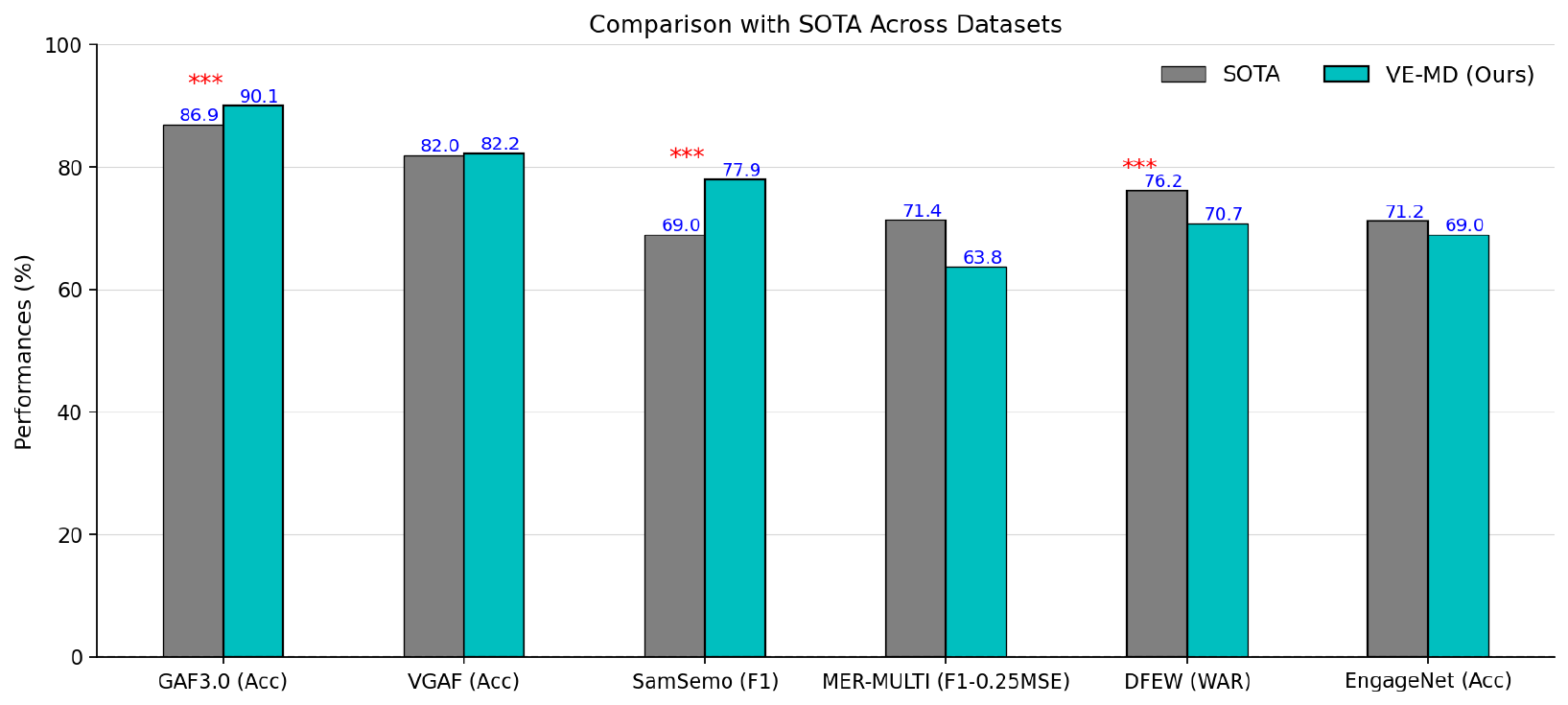}
\caption{Summary of VE-MD versus SOTA across datasets. The first two columns correspond to GER datasets. Asterisks (*) indicates a statistically significant difference versus the SOTA using a two-sided McNemar test ($p < 0.05$).}
\label{fig:sota_graph}   
\end{figure}

The SOTA result of SAMSEMO~\cite{bujnowski2024samsemo} is 69.00\% for f1-score with all three modalities: audio, video, and text. The proposed \textit{VE-MD} outperforms the SOTA with {+8.94 points} by achieving the performance 77.94\%. The corresponding configuration is Wav2Vec~2.0, Whisper, VE-MD, and Mixtral features for audio, video, and text modalities, respectively. On {MER-MULTI} by using the combined metric from MER2023~\cite{lian2023mer} challenge: $F1-0.25 \times MSE$,  \textit{VE-MD} achieves {63.80\%} compared to {71.39\%}~\cite{wang2024improving} as best performance since after the challenge. In their system, they used a multi-modal emotion recognition system with Pruning-based Network Architecture Optimization (PNAO) combined with the Attention Guided Feature (AFG) module. On EngageNet, \textit{VE-MD} reaches {68.98\%} of accuracy versus {71.24\%}~\cite{abedi2024engagement} for SOTA. Their higher result relies on explicit individual features (facial landmarks) as inputs, which is avoided in the proposed system to be aligned with the privacy claim, which avoids prior individual features in input. On DFEW, the proposed \textit{VE-MD} achieved {70.31\%} less than the SOTA  with {76.21\%}~\cite{chen2024finecliper} with the top methods leveraging large vision–language models (CLIP variants, MLLMs such as LLaVA), specialized face encoders (e.g., FaceXFormer). Noticed that, one of perspectives that could be used to improve  \textit{VE-MD} is to combine with MLLM models for scene description based on a reasoning process. A summary of these comparisons is given in the Figure~\ref{fig:sota_graph}, highlighting the comparison difference. 

\section{Conclusion}
\label{sec:conclusion}

This paper introduced the Variational Encoder--Multi-Decoder (VE-MD) framework for group emotion recognition under a privacy-aware functional design. Rather than relying on identity recognition, person tracking, or per-person emotion prediction as output objectives, VE-MD is designed for aggregate group-level affect inference while operating directly on full visual scenes. VE-MD combines a variational encoder with multi-task structural supervision through two complementary decoding strategies: \textit{VE-MD-PersonQuery}, a query-based structural decoder, and \textit{VE-MD-Heatmap}, a dense alternative that naturally adapts to variable group size. Beyond the architectural contribution, the experiments reveal a central task-dependent finding: GER and IER respond differently to structural information compression. For GER, optimizing the latent representation alone is often insufficient, as compressing structural information tends to attenuate interaction-related cues that are important for collective affect inference. Preserving explicit structural outputs up to the emotion decoder, therefore, leads to stronger performance. For IER, by contrast, projected structural representations behave as a beneficial denoising bottleneck and are often more effective than preserving raw structural cues.

Experiments on six in-the-wild datasets confirm both the effectiveness of the proposed framework and the relevance of this distinction. VE-MD achieves state-of-the-art performance on GAF-3.0, VGAF, and SAMSEMO, while remaining competitive on MER-MULTI, DFEW, and EngageNet. Overall, the results suggest that collective affect recognition benefits from structural guidance that preserves relational information, whereas individual emotion recognition benefits more from compact and denoised structural representations. These findings highlight a broader perspective for affective computing: the way structural information should be integrated depends on whether the target of inference is collective or individual. Future work will investigate how interaction-related cues can be encoded more directly within the shared latent representation, with the goal of reducing reliance on explicit structural outputs while preserving strong performance on group-level affect recognition.

\section{Acknowledgement}
This research was supported by the PERSYVAL Labex (ANR11-LABX-0025) and the TALISMAN project (ANR-22-CE38-0007). This work was granted access to the HPC
resources of IDRIS under the allocation 2023-AD010614233 made by GENCI. Part of the computations presented in this paper were performed using the GRICAD infrastructure (\href{https://gricad.univ-grenoble-alpes.fr/}{https://gricad.univ-grenoble-alpes.fr/}), which is supported by Grenoble research communities.

\bibliographystyle{plainnat}
\bibliography{References.bib}

\appendix

\subsection{Additional Structural Representation Analyses}
\label{sec:appendix_sr}

We further compare the \emph{raw} and \emph{projected} body-based SR representations for \textit{VE-MD-PersonQuery} on both GER and IER datasets. The results are reported in Table~\ref{tab:proj_detr_results}. For MER-MULTI and SamSemo, projection yields substantial improvements of about 5 to 8 percentage points over the raw setting, which further supports the interpretation that projection acts as a denoising bottleneck for individual emotion recognition.

\begin{figure}%
  \centering
  \includegraphics[width=0.99\linewidth]{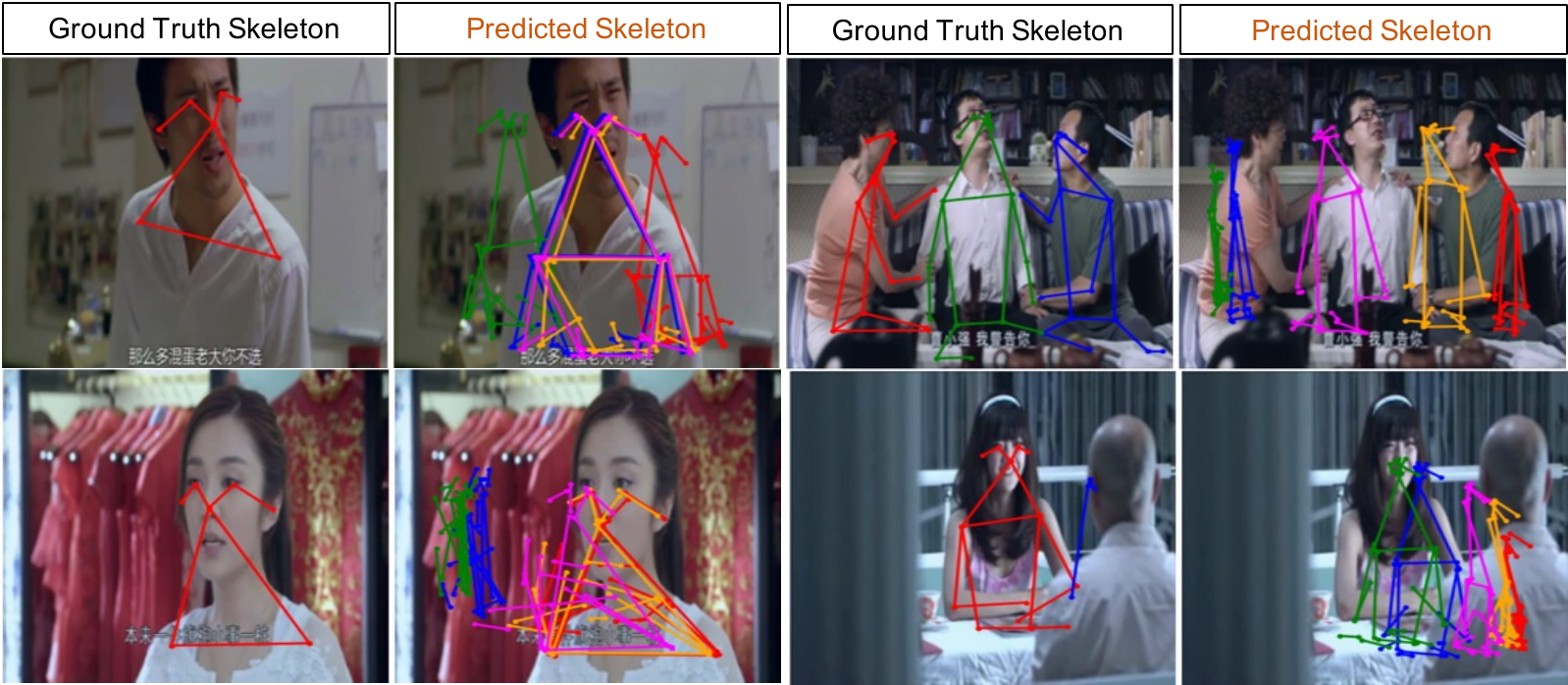}
  \caption{Qualitative examples of predicted structural representations (SR) obtained with \textit{VE-MD-PersonQuery} on MER-MULTI using a query value of 6. The model predicts more structures than those aligned with the ground truth, illustrating how noisy or redundant SR predictions can arise.}
  \label{fig:pred_SR_detr_appendix}
\end{figure}

A qualitative illustration is shown in Figure~\ref{fig:pred_SR_detr_appendix} on MER-MULTI. In the first two columns, six structural representations are predicted, although not all of them align with the ground truth. In such cases, projection can help suppress noisy or redundant structural information, improving classification. By contrast, in group emotion cases (last two columns), where the number of predicted structures is closer to the number of people present in the image, projection may discard useful interaction cues, which can harm performance.

\begin{table}[H]
\centering
\caption{Accuracy (\%) comparison between projected and non-projected body SR for \textit{VE-MD-PersonQuery} using $Q_{Max}$. Results are reported for GER and IER datasets.}
\label{tab:proj_detr_results}
\begin{autotabular}{l|cc}
\toprule
\textbf{Dataset} & \textbf{Acc. (Projected)} & \textbf{Acc. (Without projection)} \\
\midrule
GAF-3.0   & 82.65 & \textbf{84.35} \\
VGAF      & 76.24 & \textbf{78.46} \\
\midrule
MER-MULTI & \textbf{61.17} & 56.55 \\
SamSemo   & \textbf{75.29} & 67.45 \\
\bottomrule
\end{autotabular}
\end{table}

\begin{figure}%
  \centering
  \includegraphics[width=0.99\linewidth]{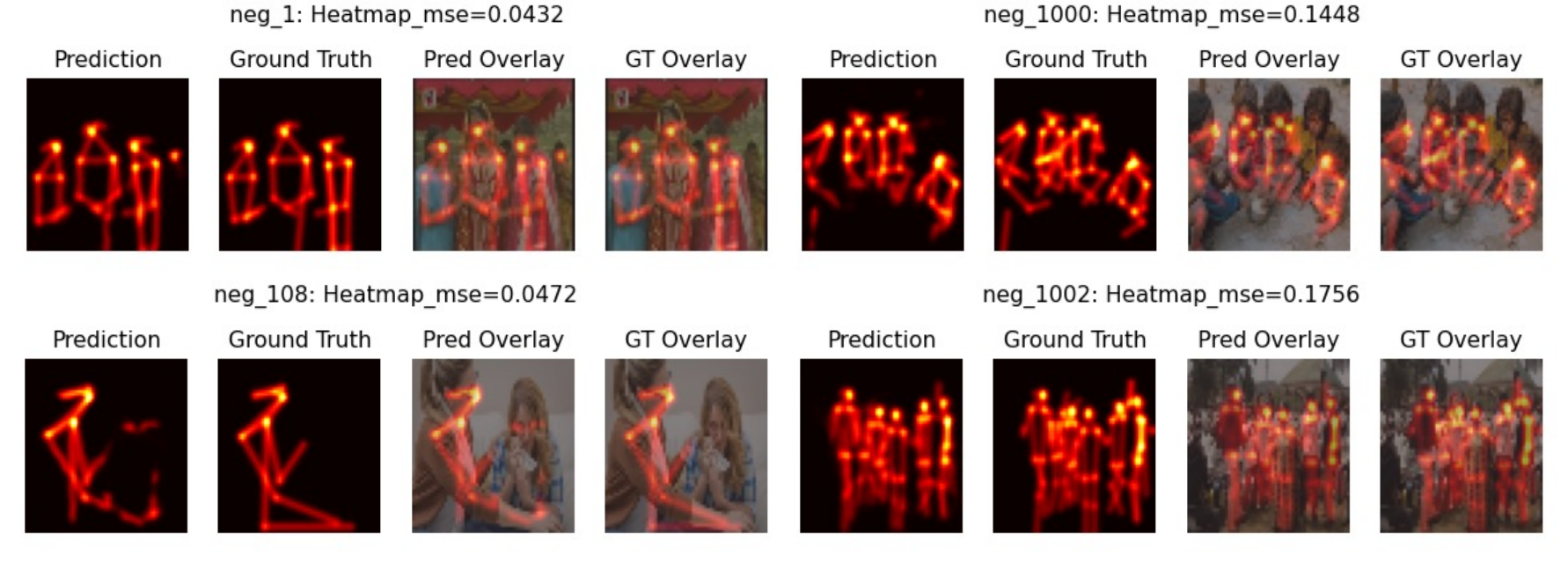}
  \includegraphics[width=0.99\linewidth]{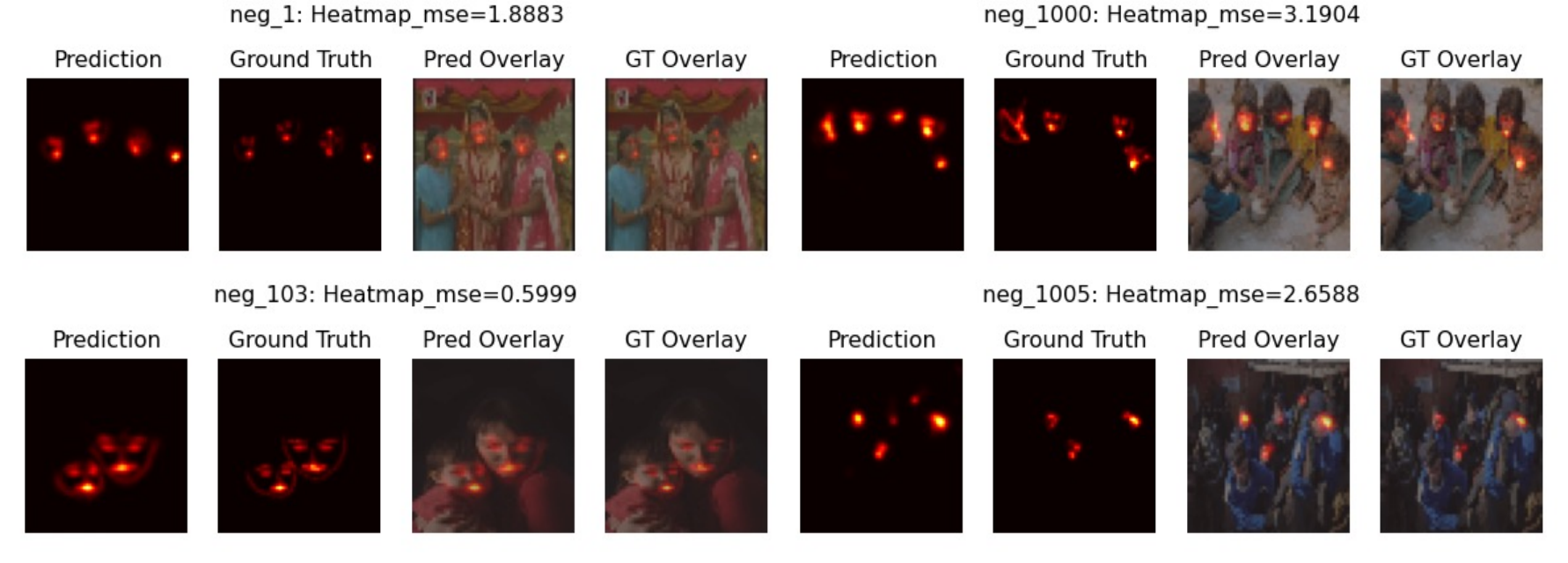}
  \caption{Predicted SR obtained with \textit{VE-MD-Heatmap} on the GAF-3.0 dataset. Unlike \textit{VE-MD-PersonQuery}, the \textit{VE-MD-Heatmap} decoder naturally adapts to the number of persons in each image. The model also often detects SR not annotated in the ground truth, reflecting higher sensitivity. For example, in image $neg\_1005$, five face SR are detected compared with three in the ground truth; in $neg\_1$, four body SR are detected compared with three labeled; and in $neg\_108$, a second person is detected despite being unlabeled.}
  \label{fig:pred_skt_heatmap_gaf3}
\end{figure}

In Figure~\ref{fig:pred_skt_heatmap_gaf3}. Unlike \textit{VE-MD-PersonQuery}, the \textit{VE-MD-Heatmap} decoder adapts to the number of visible persons in each image. It also frequently detects structural patterns not annotated in the ground truth, suggesting higher sensitivity. For example, in image $neg\_1005$, five face SR are detected although only three are annotated; in $neg\_1$, four body SR are detected although only three are labeled; and in $neg\_108$, a second person is detected despite being unlabeled.

\subsection{Additional Multimodal Fusion Results}
\label{apdx:audio_video_fusion}

\subsubsection{VGAF}
\label{apdx:vgaf_fusion}

Table~\ref{tab:multimodal_ve_md_vgaf} reports the detailed multimodal results on VGAF. The best VE-MD video accuracies are 80.81\% with face SR and 80.68\% with body SR. Late fusion with Wav2Vec~2.0 reaches 81.85\%, while combining AFG with late fusion further improves performance to 82.25\%, which is the best overall result. The two-audio-encoder setup does not exceed this score, suggesting that acoustic cues, together with AFG, provide the most useful complement for this dataset.

\begin{table}[H]
  \footnotesize
  \centering
  \caption{Detailed multimodal accuracy (\%) on VGAF.}
  \label{tab:multimodal_ve_md_vgaf}
  \begin{autotabular}{@{}lcccc@{}}
  \toprule
  \textbf{Audio Encoder} & \textbf{Fusion Strategy} & \textbf{Acc. Audio} & \textbf{Acc. Video} & \textbf{Acc. Fusion} \\
  \midrule
  Wav2Vec~2.0 & late fusion & 58.49 & 80.55 & 81.85 \\
  Whisper & late fusion & 65.79 & 80.55 & 81.20 \\
  \midrule
  Wav2Vec~2.0 & \makecell[c]{late fusion\\+ AFG} & 58.49 & 80.55 & \textbf{82.25} \\
  Whisper & \makecell[c]{late fusion\\+ AFG} & 65.79 & 80.55 & 81.59 \\
  \midrule
  \makecell[l]{Wav2Vec~2.0\\+ Whisper} &
  \makecell[c]{late fusion\\+ cross-attention} &
  65.79 & 80.55 & 81.59 \\
  \bottomrule
  \end{autotabular}
\end{table}

\subsubsection{MER-MULTI}
\label{apdx:mermulti_fusion}

Table~\ref{tab:multimodal_ve_md_mer} reports the detailed multimodal results on MER-MULTI. Whisper aligns best with the video branch under late fusion (63.02\%), whereas AFG does not provide further improvement. The best result, 64.23\%, is obtained by combining content and acoustic audio features (Whisper + Wav2Vec~2.0) with cross-attention, followed by late fusion with the video branch.

\begin{table}[H]
  \footnotesize
  \centering
  \caption{Detailed multimodal accuracy (\%) on MER-MULTI.}
  \label{tab:multimodal_ve_md_mer}
  \begin{autotabular}{@{}lcccc@{}}
  \toprule
  \textbf{Audio Encoder} & \textbf{Fusion Strategy} & \textbf{Acc. Audio} & \textbf{Acc. Video} & \textbf{Acc. Fusion} \\
  \midrule
  Wav2Vec~2.0 & late fusion & 44.52 & 62.38 & 60.84 \\
  Whisper & late fusion & 59.61 & 62.38 & 63.02 \\
  \midrule
  Wav2Vec~2.0 & \makecell[c]{late fusion\\+ AFG} & 44.42 & 62.38 & 60.35 \\
  Whisper & \makecell[c]{late fusion\\+ AFG} & 59.61 & 62.38 & 62.77 \\
  \midrule
  \makecell[l]{Wav2Vec~2.0\\+ Whisper} &
  \makecell[c]{late fusion\\+ cross-attention} &
  63.75 & 62.38 & \textbf{64.23} \\
  \bottomrule
  \end{autotabular}
\end{table}

\subsubsection{SamSemo}
\label{apdx:samsemo_fusion}

Table~\ref{tab:multimodal_ve_md_samsemo_1} reports the detailed multimodal results on SamSemo. Whisper again provides the strongest pairing with the video branch under late fusion, reaching 76.21\%, and 76.31\% when AFG is added. The best overall result is obtained by combining Wav2Vec~2.0 and Whisper with cross-attention, followed by late fusion with the video branch, yielding 78.07\% accuracy and 76.68\% weighted F1-score.

\begin{table}[H]
  \footnotesize
  \centering
  \caption{Detailed multimodal accuracy (\%) on SamSemo.}
  \label{tab:multimodal_ve_md_samsemo_1}
  \begin{autotabular}{@{}lcccc@{}}
  \toprule
  \textbf{Audio Encoder} & \textbf{Fusion Strategy} & \textbf{Acc. Audio} & \textbf{Acc. Video} & \textbf{Acc. Fusion} \\
  \midrule
  Wav2Vec~2.0 & late fusion & 61.68 & 75.31 & 74.97 \\
  Whisper & late fusion & 64.44 & 75.31 & 76.21 \\
  \midrule
  Wav2Vec~2.0 & \makecell[c]{late fusion\\+ AFG} & 61.68 & 75.31 & 75.07 \\
  Whisper & \makecell[c]{late fusion\\+ AFG} & 64.44 & 75.31 & 76.31 \\
  \midrule
  \makecell[l]{Wav2Vec~2.0\\+ Whisper} &
  \makecell[c]{cross-attention\\+ late fusion} &
  74.78 & 75.31 & \textbf{78.07} \\
  \bottomrule
  \end{autotabular}
\end{table}

\subsection{Architectural Details}
\label{apdx:architecture_details}

\subsubsection{VE-MD-Heatmap: UNet-Upsample}
\label{apdx:unet_upsample}

Table~\ref{tab:unetres} provides the architectural details of the UNet-based upsampling module used in \textit{VE-MD-Heatmap}.

\begin{table}[H]
\centering
\caption{UNetResUpsample architecture. Here, $d$ denotes the latent-space dimension and $C_{\text{out}}$ the output channel dimension. BatchNorm+ReLU follow each $3{\times}3$ convolution inside the \texttt{dec} blocks. \texttt{ResUp} upsamples by $\times2$ and maps $2048 \rightarrow 512$ channels.}
\label{tab:unetres}
\begin{autotabular}{|l|l|l|l|l|l|}
\hline
Stage & Operation & \makecell[c]{Channels\\(in$\to$out)} & Kernel / Stride / Pad & Output size & $H_0 \times W_0$ \\
\hline
\texttt{up5}   & Conv2d & $d \to 2048$ & $1{\times}1$ / $1$ / $0$ & $(H_0,\,W_0)$ & $7{\times}7$ \\
\hline
\texttt{up4}   & ResUp ($\uparrow\times2$) & $2048 \to 512$ & -- & $(2H_0,\,2W_0)$ & $14{\times}14$ \\
\hline
\texttt{dec4}  & \makecell[c]{(Conv\\+ BN\\+ ReLU)$\times2$} & $512 \to 512$ & $3{\times}3$ / $1$ / $1$ & $(2H_0,\,2W_0)$ & $14{\times}14$ \\
\hline
\texttt{up3}   & ConvT2d & $512 \to 256$ & $2{\times}2$ / $2$ / $0$ & $(4H_0,\,4W_0)$ & $28{\times}28$ \\
\hline
\texttt{dec3}  & \makecell[c]{(Conv\\+ BN\\+ ReLU)$\times2$} & $256 \to 256$ & $3{\times}3$ / $1$ / $1$ & $(4H_0,\,4W_0)$ & $28{\times}28$ \\
\hline
\texttt{up2}   & ConvT2d & $256 \to 128$ & $2{\times}2$ / $2$ / $0$ & $(8H_0,\,8W_0)$ & $56{\times}56$ \\
\hline
\texttt{dec2}  & \makecell[c]{(Conv\\+ BN\\+ ReLU)$\times2$} & $128 \to 128$ & $3{\times}3$ / $1$ / $1$ & $(8H_0,\,8W_0)$ & $56{\times}56$ \\
\hline
\texttt{up1}   & ConvT2d & $128 \to 128$ & $1{\times}1$ / $1$ / $0$ & $(8H_0,\,8W_0)$ & $56{\times}56$ \\
\hline
\texttt{dec1}  & \makecell[c]{(Conv\\+ BN\\+ ReLU)$\times2$} & $128 \to 128$ & $3{\times}3$ / $1$ / $1$ & $(8H_0,\,8W_0)$ & $56{\times}56$ \\
\hline
\texttt{final} & Conv2d & $128 \to C_{\text{out}}$ & $1{\times}1$ / $1$ / $0$ & $(8H_0,\,8W_0)$ & $56{\times}56$ \\
\hline
\end{autotabular}
\end{table}

\subsection{Extended Comparisons with the State of the Art}
\label{apdx:sota}

This appendix provides detailed comparisons with the state of the art for both GER and IER benchmarks. These extended tables complement the summary analysis reported in Section~\ref{sec:comparison_sota} and help position VE-MD more precisely with respect to prior methods.

\subsubsection{GER Datasets}

The results on GER benchmarks show that VE-MD is particularly effective for collective affect modeling. On both GAF-3.0 and VGAF, the proposed framework achieves the best reported performance, confirming that preserving structural and interaction-related cues is especially beneficial for group emotion recognition. In addition, the multimodal extension on VGAF further strengthens the competitiveness of the framework.

On GAF-3.0, both \textit{VE-MD-Heatmap} and \textit{VE-MD-PersonQuery} outperform previous image-based approaches, with \textit{VE-MD-PersonQuery} reaching the best overall accuracy of 90.06\%. On VGAF, the video-only \textit{VE-MD} result already exceeds prior video-only methods, while the multimodal extension with Wav2Vec~2.0 further improves performance to 82.25\%, establishing the best result among the compared approaches.

\begin{table}[H]
  \footnotesize
  \centering
  \caption{Detailed comparison with the state of the art on GER datasets.}
  \label{tab:ger_sota}
  \begin{autotabular}{@{}lccccc@{}}
    \toprule
    \textbf{Dataset} & \textbf{Year} & \textbf{Modalities} & \textbf{Methodology} & \textbf{Acc. [\%]} & \textbf{Rank} \\
    \midrule
    \multirow{4}[2]{*}{\begin{sideways}GAF-3.0\end{sideways}} 
    & 2018 & Img & \makecell[c]{CAN, ResNet,\\SE-Net~\cite{wang2018cascade}} & 86.90 & 4 \\
    & 2025 & Img & PSMF~\cite{huangpsmf2025} & 83.58 & 3 \\
    & 2026 & Img & VE-MD-Heatmap \textit{(ours)} & 89.92 & 2 \\
    & 2026 & Img & VE-MD-PersonQuery \textit{(ours)} & \textbf{90.06} & \textbf{1} \\
    \midrule
    \multirow{5}[2]{*}{\begin{sideways}VGAF\end{sideways}}
    & 2023 & V & \makecell[c]{ViT + Synthetic data~\cite{augusma2023multimodal}} & 79.24 & 4 \\
    & 2024 & V & \makecell[c]{TimeSformer,\\YOLOv8~\cite{kumar2020object}} & 73.10 & 5 \\
    & 2024 & A,V & \makecell[c]{Streams-fusion\\(TimeSformer, Wav2Vec~2.0,\\YOLOv8)~\cite{kumar2020object}} & 81.98 & 2 \\
    & 2026 & V & VE-MD \textit{(ours)} & 80.81 & 3 \\
    & 2026 & A,V & \makecell[c]{VE-MD + Wav2Vec~2.0 \textit{(ours)}} & \textbf{82.25} & \textbf{1} \\
    \bottomrule
  \end{autotabular}
\end{table}

\subsubsection{IER Datasets}

For IER benchmarks, the results are more mixed. VE-MD achieves state-of-the-art performance on SamSemo in the multimodal setting, showing that the framework generalizes beyond GER when combined with complementary modalities. On MER-MULTI, EngageNet, and DFEW, VE-MD remains competitive but does not surpass the strongest specialized systems. This is consistent with the main discussion of the paper: the proposed framework provides its clearest advantage when relational structural cues play a central role.

SamSemo is the strongest IER benchmark for VE-MD in the present study. When combined with audio and text features, the proposed framework achieves the best overall weighted F1-score, clearly surpassing the original benchmark baselines. This result indicates that VE-MD can transfer effectively beyond GER when multimodal information is available.

\begin{table}[H]
  \footnotesize
  \centering
  \caption{Detailed comparison with the state of the art on SamSemo.}
  \label{tab:samsemo_sota}
  \begin{autotabular}{lcccc}
    \toprule
    Year & Modalities & Methodology & \textbf{F1-score [\%]} & \textbf{Rank} \\
    \midrule
    2024 & A & E2E~\cite{bujnowski2024samsemo} & 61.10 & 9 \\
    2024 & T & E2E~\cite{bujnowski2024samsemo} & 63.00 & 8 \\
    2024 & V & E2E~\cite{bujnowski2024samsemo} & 68.20 & 7 \\
    2024 & A,V,T & E2E~\cite{bujnowski2024samsemo} & 69.00 & 6 \\
    \midrule
    2026 & A & \makecell[c]{Wav2Vec~2.0 + Whisper \textit{(ours)}} & 74.78 & 5 \\
    2026 & A,V & \makecell[c]{Whisper + VE-MD \textit{(ours)}} & 74.97 & 4 \\
    2026 & A,V & \makecell[c]{Wav2Vec~2.0 + Whisper + VE-MD \textit{(ours)}} & 76.68 & 3 \\
    2026 & A,V,T & \makecell[c]{Wav2Vec~2.0 + Whisper + VE-MD\\+ Llama2 Feat. \textit{(ours)}} & 77.71 & 2 \\
    2026 & A,V,T & \makecell[c]{Wav2Vec~2.0 + Whisper + VE-MD\\+ Mixtral Feat. \textit{(ours)}} & \textbf{77.94} & \textbf{1} \\
    \bottomrule
  \end{autotabular}
\end{table}

On MER-MULTI, VE-MD improves clearly over the original baseline but remains below the strongest recent multimodal systems. This suggests that, although structural supervision is beneficial, the current framework is not yet sufficient to match the most optimized fusion-oriented approaches on this benchmark.

\begin{table}[H]
  \footnotesize
  \centering
  \caption{Detailed comparison with the state of the art on MER-MULTI.}
  \label{tab:mermulti_sota}
  \begin{autotabular}{@{}lcccc@{}}
    \toprule
    Year & Modalities & Methodology & \makecell[c]{F1-0.25MSE [\%]} & Rank \\
    \midrule
    2023 & A,V & Baseline~\cite{lian2023mer} & 56.00 & 5 \\
    2023 & A,V & \makecell[c]{JDEV, HuBERT~\cite{wang2023hierarchical}} & 68.46 & 3 \\
    2023 & A,V & \makecell[c]{Weighted blending of\\supervision signals~\cite{zong2023building}} & 70.05 & 2 \\
    2024 & A,V & PNAO~\cite{wang2024improving} & \textbf{71.39} & \textbf{1} \\
    \midrule
    2026 & A,V & \makecell[c]{Wav2Vec~2.0 + Whisper\\+ VE-MD \textit{(ours)}} & 63.80 & 4 \\
    \bottomrule
  \end{autotabular}
\end{table}

On EngageNet, VE-MD remains competitive and ranks close to the best recent methods, but does not exceed the landmark-based ST-GCN approach. This result is still encouraging, given that VE-MD does not rely on explicit individual facial landmark features as direct inputs for the final prediction.

\begin{table}[H]
  \footnotesize
  \centering
  \caption{Detailed comparison with the state of the art on EngageNet.}
  \label{tab:engagenet_sota}
  \begin{autotabular}{lccc}
    \toprule
    Year & Methodology & \textbf{Acc. [\%]} & \textbf{Rank} \\
    \midrule
    2023 & Transformer~\cite{singh2023have} & 55.45 & 7 \\
    2023 & Transformer~\cite{singh2023have} & 64.45 & 6 \\
    2023 & Transformer~\cite{singh2023have} & 69.10 & 2 \\
    2024 & GLAMOR-Net~\cite{anand2024exceda} & 68.72 & 5 \\
    2024 & TCCT-Net~\cite{Vedernikov_2024_CVPR} & 68.91 & 4 \\
    2024 & ST-GCN~\cite{abedi2024engagement} & \textbf{71.24} & \textbf{1} \\
    \midrule
    2026 & VE-MD \textit{(ours)} & 68.98 & 3 \\
    \bottomrule
  \end{autotabular}
\end{table}

On DFEW, both VE-MD variants outperform several earlier baselines, but remain below the strongest recent CLIP- and MLLM-based methods. These results indicate that VE-MD is competitive on dynamic facial expression recognition, although it does not yet match the performance of large vision-language systems specifically optimized for this task.

\begin{table}[H]
  \footnotesize
  \centering
  \caption{Detailed comparison with the state of the art on DFEW.}
  \label{tab:dfew_sota}
  \begin{autotabular}{@{}lccc@{}}
    \toprule
    Year & Methodology & \textbf{WAR [\%]} & \textbf{Rank} \\
    \midrule
    2024 & \makecell[c]{EmoCLIP, CLIP-ViT-B/32~\cite{foteinopoulou2024emoclip}} & 62.12 & 8 \\
    2024 & \makecell[c]{OUS, CLIP~\cite{mai2024ous}} & 68.85 & 7 \\
    2024 & \makecell[c]{LSGT, ResNet-18~\cite{wang2024joint}} & 72.34 & 4 \\
    2024 & \makecell[c]{UMBEnet, CLIP~\cite{mai2024all}} & 73.93 & 3 \\
    2024 & \makecell[c]{Align-DFER, CLIP-ViT-L/14, MLLM~\cite{tao2024align}} & 74.20 & 2 \\
    2024 & \makecell[c]{FineCLIPER, CLIP-ViT-L/16, LLaVA, MLLM~\cite{chen2024finecliper}} & \textbf{76.21} & \textbf{1} \\
    \midrule
    2026 & VE-MD-PersonQuery \textit{(ours)} & 69.98 & 6 \\
    2026 & VE-MD-Heatmap \textit{(ours)} & 70.31 & 5 \\
    \bottomrule
  \end{autotabular}
\end{table}

\end{document}